\documentclass{applemlr}

\usepackage{hyperref}
\usepackage{url}
\usepackage{amsthm}
\usepackage{hyperref}

\usepackage{fontawesome5}
\usepackage[smartEllipses]{markdown}

\usepackage[most]{tcolorbox}
\usepackage{wrapfig}
\usepackage{multirow}
\usepackage{graphicx,float}
\usepackage{threeparttable}
\usepackage[ruled,vlined]{algorithm2e}
\usepackage{colortbl}
\usepackage{color}
\usepackage{multirow}
\usepackage{tabularx}
\usepackage{float}
\usepackage{graphicx}
\usepackage{booktabs}
\usepackage{arydshln}
\usepackage{enumitem}
\usepackage{wrapfig}
\usepackage{caption}
\usepackage{graphicx}
\usepackage{makecell}
\usepackage{float} 
\usepackage{mathtools}
\usepackage{bbding}
\usepackage{makecell}
\usepackage{tabularx}
\usepackage{amssymb,mathrsfs,amsmath}
\usepackage{pifont}
\usepackage{bbm}
\usepackage{hhline}
\usepackage{boldline}
\usepackage{lmodern}

\usepackage{listings}

\newcommand{\ourbench}{{\fontfamily{lmtt}\selectfont \textbf{TiViBench}}\xspace}
\newcommand{\ourmethod}{{\fontfamily{lmtt}\selectfont \textbf{VideoTPO}}\xspace}

\newcommand{\hlfirst}[1]{\colorbox{cyan!24}{#1}}  
\newcommand{\hlsecond}[1]{\colorbox{cyan!10}{#1}}

\newcommand{\hllfirst}[1]{\colorbox{Goldenrod!42}{#1}}  
\newcommand{\hllsecond}[1]{\colorbox{Goldenrod!20}{#1}}

\definecolor{newgray}{gray}{0.4}

\usepackage[table,xcdraw,usenames,dvipsnames]{xcolor}
\usepackage{cleveref}
\SetKwInOut{Input}{Input}\SetKwInOut{Output}{Output}
\usepackage{twemojis}

\hypersetup{
    colorlinks=true,
    linkcolor=magenta,
    citecolor=cyan,
    filecolor=magenta,      
    urlcolor=magenta,
    }

\title{\ourbench: Benchmarking Think-in-Video Reasoning for Video Generative Models}

\author[1,2\dagger]{Harold Haodong Chen}
\author[3\dagger]{Disen Lan}
\author[2\dagger]{Wen-Jie Shu}
\author[4]{Qingyang Liu}
\author[1]{Zihan Wang}
\author[1,7]{Sirui Chen}
\author[1]{Wenkai Cheng}
\author[1,2,7]{Kanghao Chen}
\author[1]{Hongfei Zhang}
\author[1,2,7]{Zixin Zhang}
\author[5]{Rongjin Guo}
\author[6\faEnvelope]{Yu Cheng}
\author[1,2\faEnvelope]{Ying-Cong Chen}
\affiliation[1]{HKUST(GZ)}
\affiliation[2]{HKUST}
\affiliation[3]{FDU}
\affiliation[4]{SJTU}
\affiliation[5]{CityUHK}
\affiliation[6]{CUHK}
\affiliation[7]{Knowin}
\contribution[\dagger]{Equal Contribution}
\contribution[\faEnvelope]{Corresponding Author}
\abstract{The rapid evolution of video generative models has shifted their focus from producing visually plausible outputs to tackling tasks requiring physical plausibility and logical consistency. However, despite recent breakthroughs such as Veo 3's chain-of-frames reasoning, it remains unclear whether these models can exhibit reasoning capabilities similar to large language models (LLMs). Existing benchmarks predominantly evaluate visual fidelity and temporal coherence, failing to capture higher-order reasoning abilities.
To bridge this gap, we propose \textbf{\ourbench}, a hierarchical benchmark specifically designed to evaluate the reasoning capabilities of image-to-video (I2V) generation models. \ourbench systematically assesses reasoning across four dimensions: i) \textbf{Structural Reasoning \& Search}, ii) \textbf{Spatial \& Visual Pattern Reasoning}, iii) \textbf{Symbolic \& Logical Reasoning}, and iv) \textbf{Action Planning \& Task Execution}, spanning $24$ diverse task scenarios across $3$ difficulty levels. Through extensive evaluations, we show that commercial models (\textit{e.g.}, Sora 2, Veo 3.1) demonstrate stronger reasoning potential, while open-source models reveal untapped potential that remains hindered by limited training scale and data diversity.
To further unlock this potential, we introduce \textbf{\ourmethod}, a simple yet effective test-time strategy inspired by preference optimization. By performing LLM self-analysis on generated candidates to identify strengths and weaknesses, \ourmethod significantly enhances reasoning performance without requiring additional training, data, or reward models. Together, \ourbench and \ourmethod pave the way for evaluating and advancing reasoning in video generation models, setting a foundation for future research in this emerging field.}

\metadata[{\faGlobe\ Project}]{\url{https://haroldchen19.github.io/TiViBench-Page/}}
\metadata[{\faGithub\ Github}]{\url{https://github.com/EnVision-Research/TiViBench}}

\begin{document}
\maketitle

\begin{figure*}[!h]
\centering
% \vspace{-0.4em}
\includegraphics[width=\linewidth]{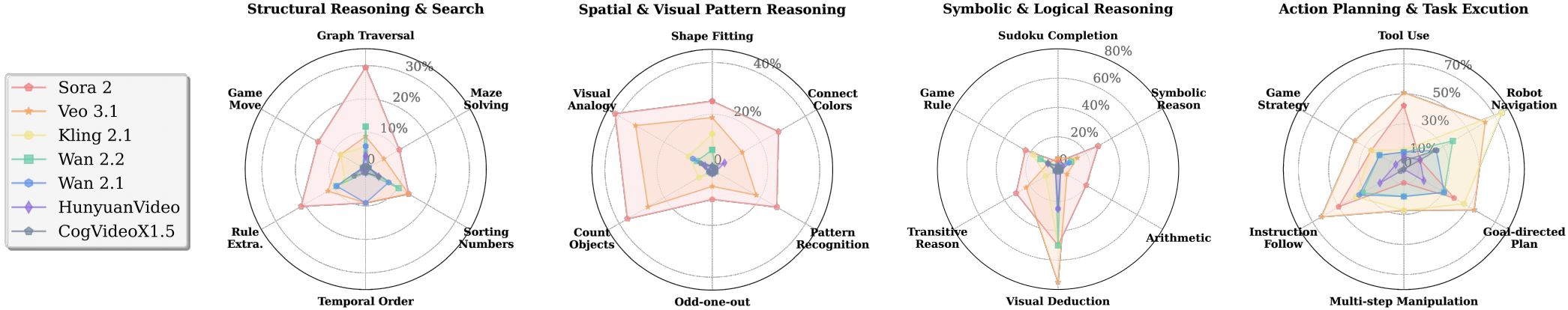}
\vspace{-1.9em}
\caption{Pass@$1$ performance overview on \ourbench across $24$ tasks within $4$ dimensions.}
\label{fig:figure3}
\vspace{-0.4em}
\end{figure*}

\section{Introduction}
\label{sec:intro}

\begin{wrapfigure}{r}{0.46\textwidth}
\vspace{-0.2em}
 \centering
 \includegraphics[width=\linewidth]{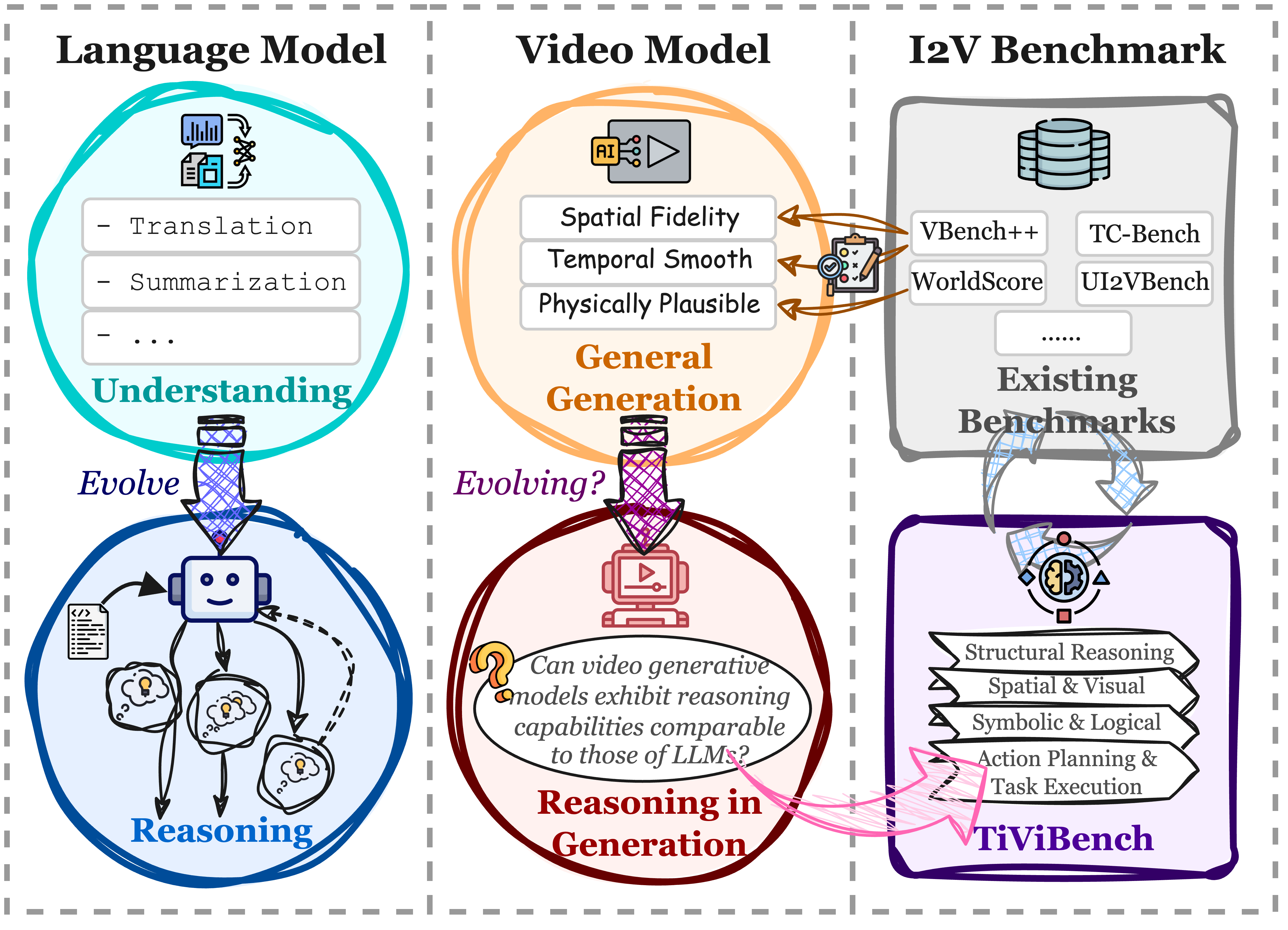}
  % \fbox{\rule[-.5cm]{0cm}{5cm} \rule[-.5cm]{5cm}{0cm}}
  \vspace{-1.9em}
  \caption{(\textbf{\textit{Left}}) Language models have evolved from basic understanding tasks to advanced reasoning capabilities. (\textbf{\textit{Middle}}) Can video generative models exhibit reasoning capabilities comparable to those of LLMs? (\textbf{\textit{Right}}) Existing I2V benchmarks focus on general generation capabilities (\textit{e.g.}, spatial fidelity, temporal smoothness), while our \ourbench complements these by introducing a reasoning-oriented benchmark, enabling comprehensive evaluation across both general and reasoning abilities.}
  \vspace{-1.6em}
  % \vspace{-4.4em}
  \label{fig:figure2}
\end{wrapfigure}

The rapid development of large language models (LLMs) \citep{achiam2023gpt, brown2020language, bai2023qwen, guo2025deepseek} has fundamentally transformed the field of artificial intelligence, pushing the boundaries of what machines can achieve in both understanding and reasoning. Initially excelling at tasks requiring basic comprehension, LLMs have evolved to tackle complex reasoning problems step-by-step \citep{hao2025can, gu2025thinkmorph}, as shown in Figure~\ref{fig:figure2} (\textit{Left}). Similarly, visual generative models \citep{rombach2022high, he2022latent, batifol2025flux, wu2025qwen, wan2025, chen2025hierarchical} have transitioned from producing
visually plausible outputs to tackling more sophisticated tasks that require physical plausibility and logical consistency. Among these, video generation has emerged as a particularly promising paradigm, with a wide range of applications, \textit{e.g.}, vision-language action \citep{chi2025wow, li2025vla, zhang2025gevrm} and novel view synthesis \citep{you2025nvssolver, zhou2025stable, voleti2024sv3d}.

A natural question thus arises: \textit{can video generative models exhibit reasoning capabilities comparable to those of LLMs?} The recent breakthrough of Veo 3 \citep{wiedemer2025video} has hinted at this possibility by introducing the concept of "chain-of-frames" reasoning in image-to-video (I2V) generation, highlighting the possibility of leveraging video frame generation as a medium for step-by-step visual reasoning. 
% By generating sequences of frames that capture temporal dynamics and spatial transformations, video models may possess the potential to reason visually across time and space. 
This raises the intriguing prospect of a "GPT moment" for video generation models: one in which they transcend their current focus on visual fidelity and become general-purpose vision foundation models capable of solving complex reasoning tasks. However, despite the promising advancements, existing benchmarks \citep{duan2025worldscore, feng2024tc, mengtowards, huang2024vbench++} for video generation fail to evaluate such reasoning abilities adequately (see Figure~\ref{fig:figure2} (\textit{Right})). Current evaluations predominantly focus on visual fidelity, temporal smoothness, physical plausibility, and adherence to input prompts, which, while essential, fail to capture higher-order reasoning abilities. This gap motivates the need for a new, complementary benchmark that can rigorously evaluate the reasoning potential of video models, paving the way for future research.

\begin{figure*}[!t]
\centering
% \vspace{-0.4em}
\includegraphics[width=\linewidth]{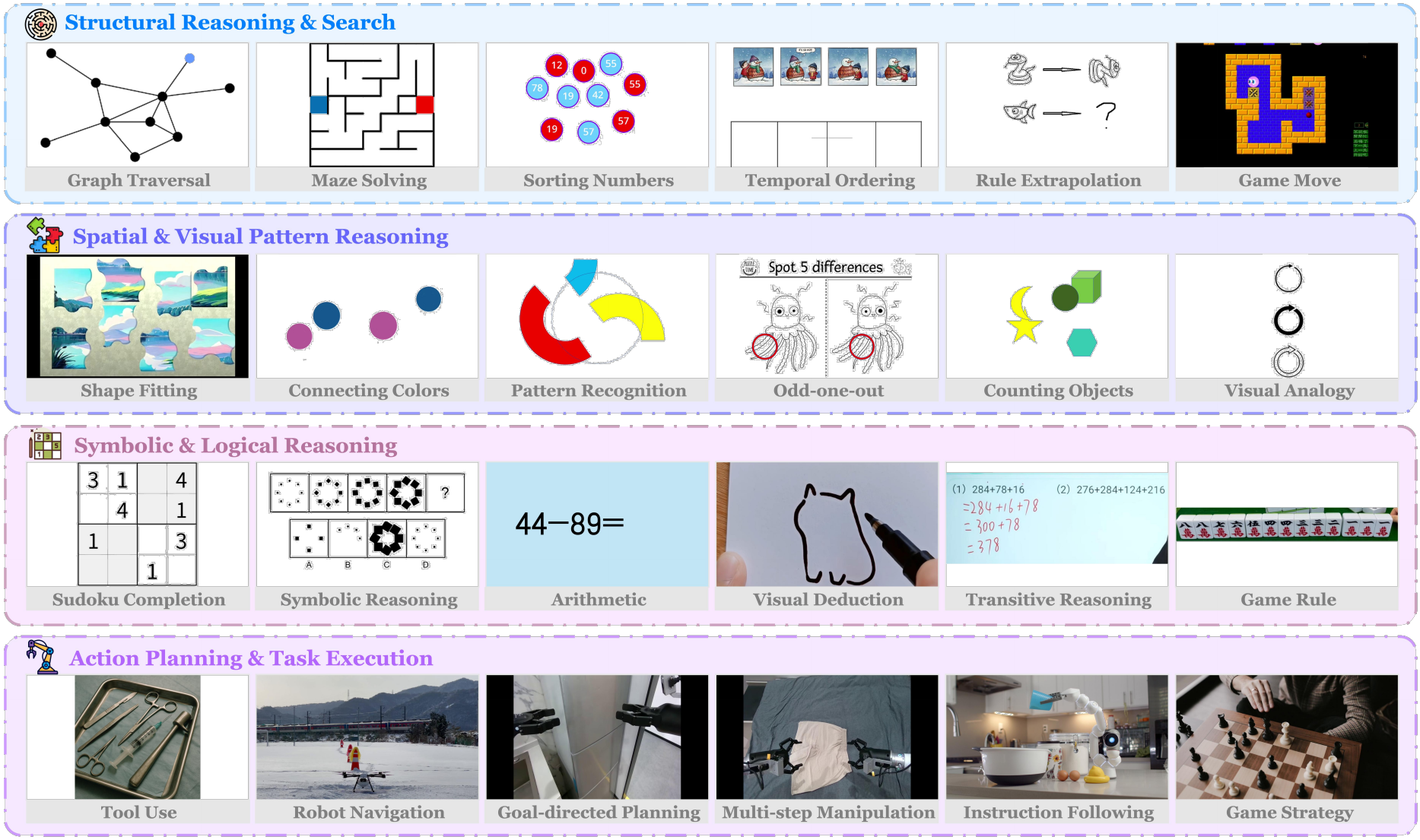}
\vspace{-1.9em}
\caption{Overview of \textbf{\ourbench}. \ourbench represents an image-to-video (I2V) benchmark tailored to comprehensively evaluate the emerging visual reasoning capabilities across four key categories: (\textbf{\textit{1st}}) Structural Reasoning \& Search, (\textbf{\textit{2nd}}) Spatial \& Visual Pattern Reasoning, (\textbf{\textit{3rd}}) Symbolic \& Logical Reasoning, and (\textbf{\textit{4th}}) Action Planning \& Task Execution. Each category encompasses six diverse tasks to challenge video generative models to perform complex reasoning beyond general generation.}
\label{fig:teaser}
\vspace{-0.6em}
\end{figure*}

% \begin{figure}[!t]
% \centering
% % \vspace{-0.4em}
% \includegraphics[width=\linewidth]{figures/Figure2_new.png}
% \vspace{-2em}
% \caption{(\textbf{\textit{Left}}) Language models have evolved from basic understanding tasks to advanced reasoning capabilities. (\textbf{\textit{Middle}}) Can video generative models exhibit reasoning capabilities comparable to those of LLMs? (\textbf{\textit{Right}}) Existing I2V benchmarks focus on general generation capabilities (\textit{e.g.}, spatial fidelity, temporal smoothness), while our \ourbench complements these by introducing a reasoning-oriented benchmark, enabling comprehensive evaluation across both general and reasoning abilities.
% }
% \label{fig:figure2}
% \vspace{-1em}
% \end{figure}

In this work, we propose \textbf{\ourbench}, a hierarchical benchmark designed specifically to evaluate the reasoning capabilities of I2V generation. Building on Veo 3’s \citep{wiedemer2025video} testing tasks like graph traversal and maze solving, we expand and diversify the evaluation scope to include more complex scenarios, \textit{e.g.}, strategic card game reasoning and mathematical problem solving, as demonstrated in Figure~\ref{fig:teaser}. Specifically, \ourbench offers a systematic suite structured around four key dimensions: 
\ding{182}~\textbf{\textit{Structural Reasoning \& Search}}, testing structure traversal, pathfinding, and constrained exploration;
\ding{183}~\textbf{\textit{Spatial \& Visual Pattern Reasoning}}, assessing capacities to detect, complete, or extrapolate patterns across time and space;
\ding{184}~\textbf{\textit{Symbolic \& Logical Reasoning}}, focusing on higher-order abstract reasoning tasks; and
\ding{185}~\textbf{\textit{Action Planning \& Task Execution}}, evaluating multi-step actions in a temporally coherent manner.
Each dimension comprises about $150$ evaluation samples across three hierarchical levels (\textit{i.e.}, easy, medium, hard), totally covering $24$ task scenarios.

Through extensive evaluation on \ourbench, we observe that commercial models demonstrate stronger reasoning potential compared to open-source models, as shown in Figure~\ref{fig:figure3}. However, open-source ones also exhibit potential, albeit with inconsistent performance.
To further unlock this potential, we propose a simple yet effective test-time strategy, dubbed \textbf{\ourmethod}.  Unlike strategies like SFT with domain-specific data, which are intuitively likely to enhance reasoning capabilities \citep{miniveo3reasoner} but require constructing large and diverse datasets with significant costs, \ourmethod avoids such overhead by operating entirely at test time. Specifically, different from conventional single-pass prompt rewriting \citep{xue2025phyt2v}, \ourmethod draws inspiration from test-time preference optimization \citep{li2025testtime} in LLMs. By leveraging multi-pass generation and aligning candidates through preference alignment, enabling more fine-grained and accurate prompt optimization, \ourmethod serves as both a complementary method to \ourbench and a practical solution for improving reasoning performance without weight updates.

To summarize, this work contributes threefold: 
% \vspace{-0.4em}
\begin{itemize}[leftmargin=*]
    \item We propose \textbf{\ourbench}, a hierarchical benchmark tailored to systematically evaluate the reasoning capabilities of video generative models, covering $4$ reasoning dimensions across $24$ diverse scenarios and $3$ difficulty levels.
    \item Through extensive experiments, we analyze the reasoning potential of $3$ commercial and $4$ open-source advanced video models. Our findings highlight the advantages of commercial models while exposing their limitations, and reveal the latent capabilities of open-source models that remain constrained by current scales.
    \item We further introduce \textbf{\ourmethod}, a simple yet effective test-time strategy that unlocks reasoning potential on-the-fly via preference optimization. It requires no additional training, data, or reward models, offering a scalable solution for improving video generation models.
\end{itemize}

\section{Related Work}
% \vspace{-0.4em}

\paragraph{Image-to-Video Generation.}

% The rapid development of generative models has expanded visual generation from images \citep{batifol2025flux, wu2025qwen, rombach2022high} to videos \citep{wan2025, openai_sora2_2025, yang2024cogvideox, kong2024hunyuanvideo}. As a subfield of video generation, I2V generation has gained attention for its ability to produce more personalized outputs compared to text-to-video (T2V) models \citep{ni2023conditional, hu2024animate, karras2023dreampose}. Beyond improving general capabilities, I2V has also served as a testing ground for pioneering concepts in video generation. For example, early studies on physical plausibility in video generation explored scenarios like ball collisions \citep{liu2024physgen, yang2025vlipp} and free fall \citep{li2025pisa} within I2V settings before extending these ideas to T2V models \citep{yuan2025newtongen, chen2025hierarchical, wang2025wisa}.
Recent advancements in generative models have extended visual generation from images \citep{batifol2025flux, wu2025qwen, rombach2022high, wu2025lightgen} to videos \citep{wan2025, openai_sora2_2025, yang2024cogvideox, kong2024hunyuanvideo}. As a subfield, image-to-video (I2V) generation enables more personalized outputs compared to text-to-video (T2V) models \citep{ni2023conditional, hu2024animate, karras2023dreampose, shao2025finephys} and has served as a testing ground for key concepts like physical plausibility \citep{liu2024physgen, yang2025vlipp, li2025pisa}.
With general generation capabilities reaching new heights, researchers \citep{yang2024video, wiedemer2025video} have recently begun investigating whether video generative models can exhibit reasoning abilities akin to LLMs.
% A recent work, VChain, introduced the concept of chain-of-visual thoughts. However, VChain primarily enriches prompts using LLM reasoning to enhance physical plausibility, rather than investigate visual reasoning directly.
However, there is currently a lack of systematic benchmarks to evaluate reasoning capabilities in video generation models. 
% Our work addresses this gap by introducing \ourbench, a comprehensive benchmark for visual reasoning in I2V generation.

\vspace{-1em}
\paragraph{Evaluation of I2V Models.}

Early evaluations of I2V models relied on metrics like FVD \citep{heusel2017gans} on datasets like UCF101 \citep{soomro2012ucf101},
later expanded by more recent benchmarks with fine-grained dimensions (\textit{e.g.}, 10 I2V dimensions in VBench++ \citep{huang2024vbench++}). These benchmarks \citep{zhang2025ui2v, feng2024tc, huang2024vbench++, duan2025worldscore, zhang2024pia, fan2023aigcbench} provide robust standards for assessing general generation capabilities, including spatial fidelity, temporal smoothness, and physical plausibility.
However, none of these benchmarks systematically evaluate visual reasoning. 
% One concurrent work, MME-CoF \citep{guo2025video}, introduces 12 reasoning dimensions derived from specific task types. While its fine-grained design provides detailed insights, it lacks a hierarchical framework to differentiate tasks by difficulty, treating simple tasks (\textit{e.g.}, "rotation reasoning") and complex ones (\textit{e.g.}, "long-horizon causal reasoning") equally, which limits its ability to reveal nuanced model behaviors.
Recent concurrent works, such as MME-CoF \citep{guo2025video} and VideoThinkBench \citep{tong2025thinking}, have begun to explore reasoning capabilities in video generation models. MME-CoF focuses on fine-grained reasoning dimensions derived from specific task types, while VideoThinkBench evaluates video generation models on both vision-centric and text-centric tasks. These works highlight the potential of video generation models for reasoning but focus on specific task designs or general multimodal reasoning without a dedicated framework for systematically scaling task difficulty.
In contrast, we propose \ourbench, a hierarchical benchmark dedicated to visual reasoning in I2V models. \ourbench systematically evaluates models across $4$ high-level reasoning dimensions and $24$ task scenarios, each categorized into $3$ difficulty levels, offering a comprehensive and nuanced assessment of zero-shot reasoning capabilities.

% \begin{figure}[!t]
% \centering
% % \vspace{-0.4em}
% \includegraphics[width=0.9\linewidth]{figures/figure3.pdf}
% \vspace{-0.8em}
% \caption{\textcolor{red}{\textbf{PLACEHOLDER...TBD...}} Pass@$1$ performance overview across $24$ tasks on \ourbench.
% }
% \label{fig:figure3}
% \vspace{-1em}
% \end{figure}

\vspace{-1em}
\paragraph{Prompt Optimization for Video Generative Model.}

% SFT \citep{brown2020language} and reinforcement fine-tuning (RFT) \citep{shao2024deepseekmath, rafailov2023direct} are effective for enhancing specific capabilities, but they incur high costs due to additional data and training requirements.
While supervised fine-tuning (SFT) \citep{brown2020language} and reinforcement fine-tuning (RFT) \citep{shao2024deepseekmath, rafailov2023direct} enhance specific capabilities, they incur high costs due to additional data and training.
% Recent efforts, such as fine-tuning video models \citep{wan2025} on maze-solving datasets \citep{ivanitskiy2023configurable}, demonstrate scalability but conflict with \ourbench's goal of evaluating inherent reasoning potential without further training.
% To this end, test-time prompt optimization offers a lightweight alternative. Existing methods can be categorized into pre-inference \citep{wan2025, wiedemer2025video} and post-inference \citep{xue2025phyt2v} rewriting.
Test-time prompt optimization offers a lightweight alternative. Existing methods can be categorized into pre-inference \citep{wan2025, wiedemer2025video} and post-inference \citep{xue2025phyt2v} rewriting.
The former enriches prompts using LLMs for reasoning or imagination but risks deviating from user intent, while the later iteratively refines prompts based on generated results, improving outputs. However, single-pass strategies (\textit{i.e.}, generating one sample per round) limit optimization granularity. To address this, we propose \ourmethod, inspired by test-time preference optimization \citep{li2025testtime}. By generating multiple candidate videos, \ourmethod identifies both general shortcomings and model-specific preferences, enabling a more fine-grained optimization for video generative reasoning without parameter updates.

\section{\ourbench: Benchmarking Visual Reasoning Potential}

\begin{figure*}[!t]
\centering
% \vspace{-0.4em}
\includegraphics[width=\linewidth]{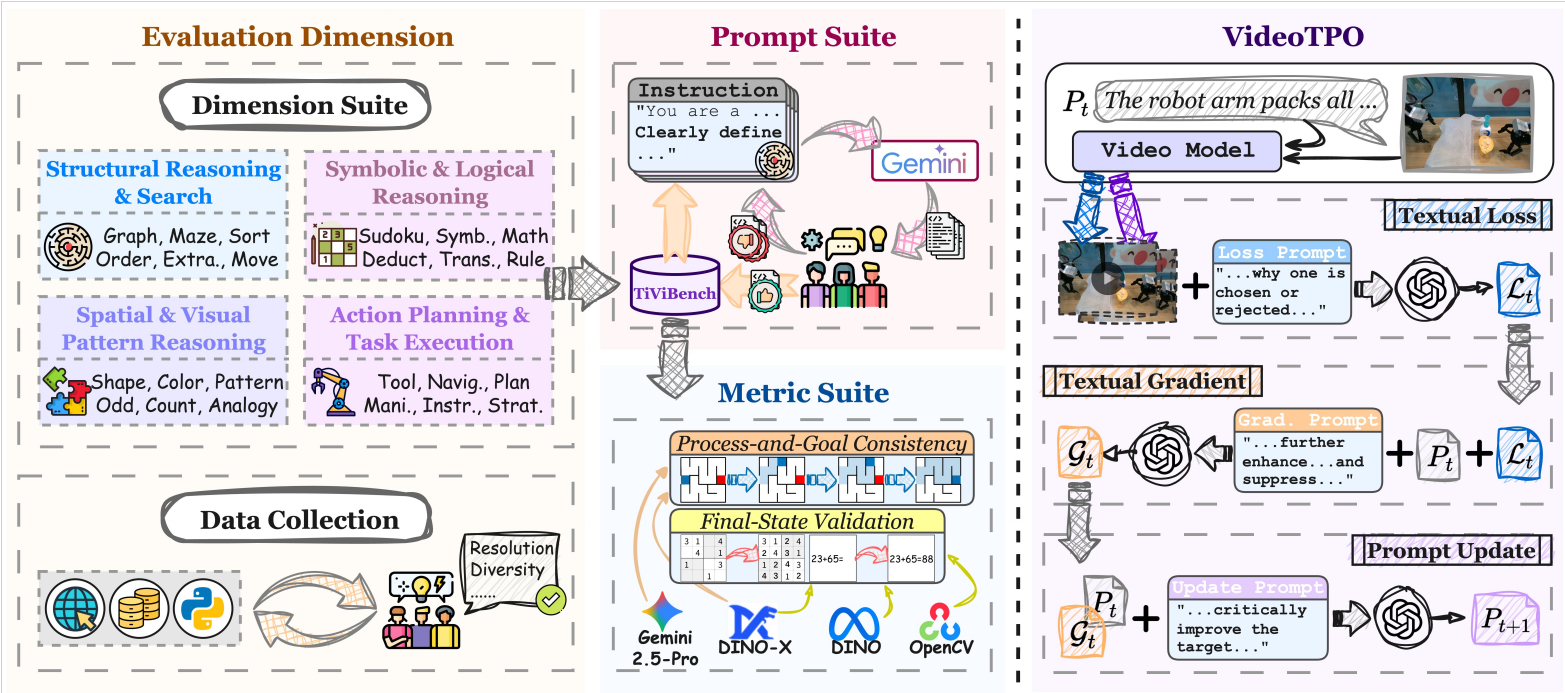}
\vspace{-2em}
\caption{Overview of our proposed (\textbf{\textit{Left}}) \ourbench benchmark and (\textbf{\textit{Right}}) \ourmethod framework. }
\label{fig:figure4}
\vspace{-0.6em}
\end{figure*}

\begin{figure*}[!t]
\centering
% \vspace{-0.4em}
\includegraphics[width=\linewidth]{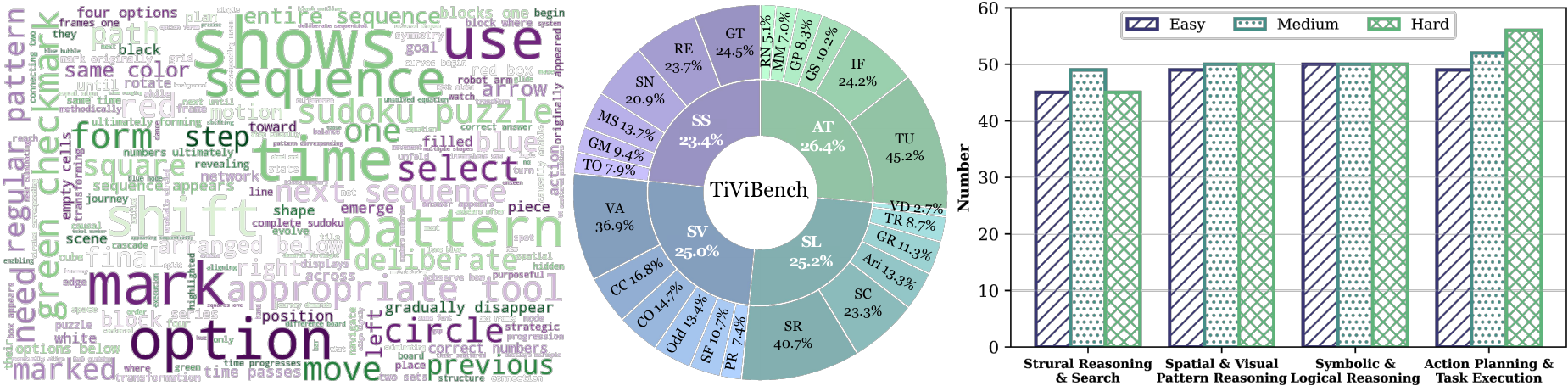}
\vspace{-1.9em}
\caption{Overview of \ourbench's statistical distributions. (\textbf{\textit{Left}}) Word distribution of prompt suites; (\textbf{\textit{Middle}}) Data distribution across $24$ tasks; and (\textbf{\textit{Right}}) Data distribution across $3$ difficulty levels. }
\label{fig:figure5}
\vspace{-0.6em}
\end{figure*}

To evaluate the visual reasoning capabilities of I2V generation, we introduce \ourbench, as shown in Figure~\ref{fig:figure4} (\textit{Left}), a comprehensive benchmark covering $4$ dimensions, $24$ task scenarios, and $595$ image-prompt samples. Each dimension is structured around $3$ difficulty levels: easy, medium, and hard. \ourbench provides a foundation for evaluating video generative reasoning. We next detail the evaluation dimension (\S\ref{sec:3.1}), prompt suite (\S\ref{sec:3.2}), and metric suite (\S\ref{sec:3.3}), with data statistics shown in Figure~\ref{fig:figure5}.

\subsection{Evaluation Dimension}
\label{sec:3.1}

To comprehensively evaluate the visual reasoning capabilities of I2V generation models, we extend and diversify the testing tasks introduced in \citep{wiedemer2025video}. Our benchmark spans four key reasoning dimensions: (\textbf{\textit{i}}) \textit{Structural Reasoning \& Search}, (\textbf{\textit{ii}}) \textit{Spatial \& Visual Pattern Reasoning}, (\textbf{\textit{iii}}) \textit{Symbolic \& Logical Reasoning}, and (\textbf{\textit{iv}}) \textit{Action Planning \& Task Execution}. Each dimension includes tasks designed to probe distinct reasoning abilities, as shown in Figure~\ref{fig:teaser}, with samples categorized into three difficulty levels.

\vspace{-1em}
\paragraph{Structural Reasoning \& Search.} This dimension focuses on a model’s ability to understand and navigate structured environments, solve constrained problems, and extrapolate patterns. Tasks in this category emphasize logical exploration, temporal coherence, and systematic problem-solving, including: \ding{192} \textit{graph traversal}, \ding{193} \textit{maze solving}, \ding{194} \textit{sorting numbers}, \ding{195} \textit{temporal ordering}, \ding{196} \textit{rule extrapolation}, and \ding{197} \textit{game move reasoning}.
% These tasks evaluate logical exploration, temporal coherence, and systematic problem-solving.

\vspace{-1em}
\paragraph{Spatial \& Visual Pattern Reasoning.} This dimension evaluates the model’s ability to recognize, manipulate, and reason about spatial relationships and visual patterns. Tasks in this category emphasize perceptual understanding and spatial transformations, including: \ding{192} \textit{shape fitting}, \ding{193} \textit{connecting colors}, \ding{194} \textit{pattern recognition}, \ding{195} \textit{odd-one-out}, \ding{196} \textit{counting objects}, and \ding{197} \textit{visual analogy}.
% These challenges probe spatial awareness, pattern recognition, and visual reasoning.

\vspace{-1em}
\paragraph{Symbolic \& Logical Reasoning.} This dimension focuses on higher-order reasoning tasks that require abstract thinking, logical inference, and symbolic manipulation. Tasks include: \ding{192} \textit{simple Sudoku completion}, \ding{193} \textit{arithmetic operations}, \ding{194} \textit{symbolic reasoning}, \ding{195} \textit{visual deduction}, \ding{196} \textit{transitive reasoning}, and \ding{197} \textit{game rule reasoning}.
% These tasks test symbolic manipulation, logical inference, and abstract reasoning.

\vspace{-1em}
\paragraph{Action Planning \& Task Execution.} This dimension evaluates the model’s ability to plan and execute multi-step actions in a temporally coherent and goal-directed manner. Tasks include: \ding{192} \textit{tool use}, \ding{193} \textit{robot navigation}, \ding{194} \textit{goal-directed planning}, \ding{195} \textit{multi-step manipulation}, \ding{196} \textit{visual instruction following}, and \ding{197} \textit{game strategy planning}.
% These tasks assess action planning and sequential task execution.

\vspace{-1em}
\paragraph{Data Collection \& Standards.} To ensure the quality and diversity of our benchmark, we collect data from three primary sources: internet data, existing datasets (\textit{e.g.}, lecture videos in Video-MMLU \citep{song2025video}, tool use images in PhysToolBench \citep{zhang2025phystoolbench}), and synthetic data created using Python scripts. Unlike previous I2V benchmarks \citep{huang2024vbench++, zhang2024pia} that primarily contain initial inference images, our focus on video data allows us to capture the initial state, process state, and the target state, enabling more reliable evaluations.
Additionally, our data collection process prioritizes quality and diversity. First, all data samples are curated to meet high-quality standards and are adapted to model input requirements, \textit{e.g.}, $720$p resolution for horizontal videos. Second, to ensure diversity, we require that samples of the same type and difficulty level differ in background, style, or format as much as possible. Finally, each sample is reviewed by at least three human annotators to ensure both quality and diversity. Details are provided in \textbf{Appendix}~\S\ref{app:dimension}.
% This rigorous process ensures that \ourbench provides a robust and reliable foundation for evaluating the reasoning capabilities of I2V models.

\subsection{Prompt Suite}
\label{sec:3.2}

Unlike the prompt style in LLM reasoning, which heavily instructs models (\textit{e.g.}, \textit{"Find the optimal path from A to B..."}), visual reasoning prompts for generative models emphasize \textbf{task subjectivity} and \textbf{narrative descriptiveness} (\textit{e.g.}, \textit{"The blue ball slides smoothly along the white path, stopping at the red point..."}). These prompts should leave room for the model to infer intermediate steps while also providing sufficient details to guide reasoning (\textit{e.g.}, \textit{"The blue ball never crosses into the black areas..."}). 

To meet these requirements, we adopt Gemini-2.5-Pro \citep{deepmind_gemini_pro_2024} as a powerful assistant for generating prompts, leveraging initial state and target state images to construct prompts that are visually grounded and reasoning-driven. Specifically, prompts are tailored to each dimension:

\vspace{-1em}
\paragraph{Structural Reasoning \& Search.}
% Prompts in this dimension test logical exploration and structured problem-solving. Key considerations include:
\ding{192} \textit{Goal Clarity}: Define start and end states without specifying the solution path; \ding{193} \textit{Implicit Rules}: Incorporate hidden constraints or rules that the model must infer; and \ding{194} \textit{Temporal Coherence}: Ensure prompts describe tasks that unfold logically over time.

\vspace{-1em}
\paragraph{Spatial \& Visual Pattern Reasoning.}
% This dimension emphasizes spatial relationships and pattern recognition. Key considerations include:
\ding{192} \textit{Visual Specificity}: Provide rich descriptions of visual elements, \textit{e.g.}, shapes, colors, and positions; \ding{193} \textit{Pattern Identification}: Encourage recognition and extension of visual patterns; and \ding{194} \textit{Open-ended Tasks}: Allow for multiple valid solutions.

\vspace{-1em}
\paragraph{Symbolic \& Logical Reasoning.}
% Prompts here focus on abstract reasoning and symbolic manipulation. Key considerations include:
\ding{192} \textit{Implicit Rule Discovery}: Avoid explicitly stating rules, letting models infer them from the prompt; \ding{193} \textit{Symbol-Visual Integration}: Combine symbolic reasoning with visual elements; and \ding{194} \textit{Logical Progression}: Ensure tasks involve clear logical sequences.

\vspace{-1em}
\paragraph{Action Planning \& Task Execution.}
% This dimension tests multi-step planning and execution. Key considerations include:
\ding{192} \textit{Goal-Oriented Descriptions}: Define the goal while leaving intermediate steps implicit; \ding{193} \textit{Multi-step Reasoning}: Encourage models to plan and execute sequential actions; and \ding{194} \textit{Causal Logic}: Ensure prompts with clear cause-and-effect relationships.

\vspace{-1em}
\paragraph{Prompt Quality Assurance.} After generating the initial prompts, we conducted rigorous manual reviews to ensure quality, clarity, and alignment with our \ourbench's goals. Specifically: (\textbf{\textit{i}}) each prompt is reviewed by three human annotators. Any prompt flagged by even one annotator as unclear or unsuitable is revised; and (\textbf{\textit{ii}}) only prompts meeting the expectations of all three annotators are adopted. Detailed information can be found in \textbf{Appendix}~\S\ref{app:prompt}.

\subsection{Metric Suite}
\label{sec:3.3}

Unlike general I2V benchmarks, visual reasoning tasks are inherently more verifiable due to explicit ground-truth information, including initial, intermediate, and target states. To evaluate reasoning capabilities effectively, we categorize metrics into two types, both focusing on correctness.

\vspace{-1em}
\paragraph{Process-and-Goal Consistency.} These tasks evaluate both the reasoning process and the final result, ensuring the generated video aligns with the expected trajectory and reaches the correct target state. For instance, in maze navigation, tools with tracking \citep{ren2024dino} can be used to track the subject across frames and validate the trajectory. 

\vspace{-1em}
\paragraph{Final-State Validation.} These tasks assess whether the generated video achieves the correct target state, with no emphasis on intermediate reasoning steps. For example, Sudoku completion can be validated by comparing the generated grid (\textit{e.g.}, via OpenCV \citep{opencv}) with the ground truth; and sequence completion can be validated by comparing extracted features (\textit{e.g.}, via DINO \citep{caron2021emerging}).
While metrics are grouped into these two categories, the validation method may vary across tasks and even within the same task type depending on the specific format, \textit{e.g.}, mathematical reasoning can be evaluated by checking the output after the equals sign for fill-in-the-blank tasks or the selected option for multiple-choice questions. Details in \textbf{Appendix}~\S\ref{app:metric}.

\section{\ourmethod: Prompt Preference Optimization On-the-Fly for Video Generative Reasoning}

Despite the rigorous prompt quality control in \ourbench to ensure compatibility with most I2V models, differences in pretraining data and architectures often lead to varying prompt preferences across models. To address this, we propose \ourmethod, a novel test-time prompt optimization strategy tailored for \ourbench, which aims to further unlock the potential of I2V models without additional tuning, as demonstrated in Figure~\ref{fig:figure4} (\textit{Right}).

Existing prompt rewriting methods are typically classified as \textbf{pre-inference} \citep{wan2025, wiedemer2025video} (\textit{i.e.}, enriching prompts by hallucinating details) and \textbf{post-inference} \citep{xue2025phyt2v} (\textit{i.e.}, modifying prompts based on the generation result). However, visual reasoning tasks are inherently more complex than general I2V tasks, requiring a more nuanced and adaptive approach. To this end, we introduce the concept of test-time preference optimization (TPO) \citep{li2025testtime} for language models, which enables finer-grained optimization by comparing preferences across multiple generated samples.
Different from TPO, which generates multiple samples (\textit{e.g.}, $4$) and relies on external reward models to rank preferences,
% we propose \textbf{test-time \textit{direct} preference optimization} (TDPO). Specifically,
our \ourmethod generates only two samples per round and tasks a VLM with self-analyzing their strengths and weaknesses. This eliminates external rewards, making \ourmethod \textit{as simple as possible to be practical}.

\vspace{-1em}
\paragraph{Textual Loss.} Given an inference image $I$ with corresponding text prompt $P_t$ at iteration $t$, the I2V model generates two candidate videos $V_t^1$ and $V_t^2$. We then assign a VLM (\textit{i.e.}, GPT-4o \citep{achiam2023gpt}) denoted as $\mathcal{M}$ to conduct self-analysis, which compares their strengths and weaknesses to produce textual critiques. The critiques highlight the advantages of the preferred video and the shortcomings of the non-preferred video, forming the textual loss:  
\setlength\abovedisplayskip{3.4pt}
\setlength\belowdisplayskip{3.4pt}
\begin{equation}
\mathcal{L}_t = \mathcal{M}(V_t^1, V_t^2, P_t),
\end{equation}
where $\mathcal{L}_t$ encapsulates qualitative feedback rather than numerical scores, enabling more interpretable optimization.

\vspace{-1em}
\paragraph{Textual Gradient.} Based on $\mathcal{L}_t$, the VLM generates actionable suggestions as textual gradient $\mathcal{G}_t$ \citep{yuksekgonul2025optimizing} to improve the prompt $P_t$. These suggestions guide the refinement of the prompt by specifying changes that better align the generated videos with the desired reasoning or visual outcomes:  
\begin{equation}
\mathcal{G}_t = \mathcal{M}(P_t, \mathcal{L}_t).
\end{equation}
The textual gradient $\mathcal{G}_t$ serves as a direct interpretation of the textual loss, ensuring the optimization remains lightweight and avoids reliance on external reward models.

\vspace{-1em}
\paragraph{Prompt Update.} The prompt $P_t$ is then updated iteratively using $\mathcal{G}_t$ to produce a refined prompt $P_{t+1}$:  
\begin{equation}
P_{t+1} = \mathcal{M}(P_t, \mathcal{G}_t).
\end{equation} 
Detailed task prompts regarding textual loss, gradient calculations, and updating can be found in \textbf{Appendix}~\S\ref{app:tpo}.
    
\section{Experiments}

In this section, we conduct extensive experiments to answer the following research questions: (\textbf{RQ1}) Do video generative models possess inherent reasoning potential? (\textbf{RQ2}) What are the primary factors contributing to reasoning failures? (\textbf{RQ3}) Can test-time optimization serve as an efficient and effective method to guide and enhance reasoning?

\subsection{Experimental Settings}

\paragraph{Models.} We conduct evaluation on \ourbench with advanced I2V models: \ding{182} \textit{Open-Source}: Wan2.2-I2V-A14B, Wan2.1-I2V-14B \citep{wan2025}, HunyuanVideo-I2V \citep{kong2024hunyuanvideo}, CogVideoX1.5-I2V \citep{yang2024cogvideox}. \ding{183} \textit{Commercial}: Veo 3.1-fast \citep{gemini_video_generation}, Sora 2 \citep{openai_sora2_2025}, and Kling 2.1 \citep{klingai2025}. We further apply our \ourmethod to Wan2.1-I2V-14B \citep{wan2025} and HunyuanVideo-I2V \citep{kong2024hunyuanvideo}, as neither includes a built-in prompt rewriter.

\vspace{-1em}
\paragraph{Evaluations.} Since \ourbench focuses on the correctness of visual reasoning, we report Pass$@1$ and Pass$@5$ for comparisons. Here, Pass$@k$ indicates the accuracy of the model in producing at least one correct output within the $k$ predictions, where we infer open-source models under multiple random seeds to ensure a more comprehensive evaluation. For commercial models, due to their strong performance and black-box nature, we report only Pass$@1$.
Following VBench++ \citep{huang2024vbench++}, we further adjust the input image resolution before inference to align with the preferences of each model, ensuring fair and optimal testing conditions.

% \vspace{-1em}
% \paragraph{Implementation Details}

\subsection{Main Results (RQ1)}

To answer \textbf{RQ1}, we present evaluation results across three difficulty levels, providing a global analysis of model performance on our \ourbench, as shown in Table~\ref{tab:pass@1} for Pass@$1$ and Table~\ref{tab:pass@5} for Pass@$5$ accuracy. Key observations are summarized as follows:

\vspace{-1em}
\paragraph{Takeaway \ding{182}: Sufficient data and scale likely contribute to clear reasoning potential.} From Table~\ref{tab:pass@1}, commercial models (\textit{e.g.}, Sora 2 and Veo 3.1) consistently outperform open-source models across all difficulty levels and reasoning dimensions. Notably, Sora 2 achieves the highest overall performance of $27.9\%$, demonstrating reasoning capabilities that remain robust even as task difficulty increases.
This suggests that reasoning ability is not an inherent limitation of generative models but rather emerges with access to sufficiently large and diverse datasets, coupled with high parameter scales and optimized architectures.
% The superior performance of commercial models highlights the importance of extensive training on reasoning-aligned data and the use of optimized architectures designed to handle complex temporal, symbolic, and logical dependencies.

\vspace{-1em}
\paragraph{Takeaway \ding{183}: Pass@5 improvements reveal the emerging reasoning potential of open-source models.} Table~\ref{tab:pass@5} shows a clear improvement in Pass@$5$ over Pass@$1$ for advanced open-source models (\textit{e.g.}, Wan2.2 and Wan2.1), indicating that they are capable of generating correct solutions, albeit inconsistently. This suggests that open-source models possess latent reasoning potential, but their unstable performance highlights limitations in the scale of their current training. Further scaling of training data, model parameters, or reasoning-specific optimization shows the necessity to realize the reasoning capability better.

\begingroup
\setlength{\tabcolsep}{0.9pt}
\begin{table*}[t]
% \vspace{-0.6em}
\renewcommand{\arraystretch}{1.5}
  \centering
  \caption{Pass$@1$ performance of $7$ advanced models on \ourbench. We highlight the \hlfirst{best} and \hlsecond{second best} results. }
  \centering
  \vspace{-0.7em}
  \scriptsize
   \begin{tabular}{l cccc cccc cccc cccc c}
     \hlineB{2.5}
     % \rowcolor{CadetBlue!20} 
     \multirow{2}{*}{Model} & \multicolumn{4}{c}{Structural \& Search} & \multicolumn{4}{c}{Spatial \& Visual Pattern} & \multicolumn{4}{c}{Symbolic \& Logical} & \multicolumn{4}{c}{Planning \& Execution} & \multirow{2}{*}{Overall}\\
     % \cline{2-5}\cline{6-10}
     \cmidrule(lr){2-5} \cmidrule(lr){6-9} \cmidrule(lr){10-13} \cmidrule(lr){14-17}
     % \rowcolor{CadetBlue!20} 
     & Easy & Med. & Hard & Over. & Easy & Med. & Hard & Over.  & Easy & Med. & Hard & Over. & Easy & Med. & Hard & Over. & \\
     \hlineB{1.5}
     \rowcolor{lightgray!20} \multicolumn{18}{c}{{\textit{Open-Source Models}}} \\
     CVX1.5 & 2.22 & 2.04 & 0.00 & 1.42 & 2.04 & 2.00 & 0.00 & 1.34 & 2.00 & 0.00 & 0.00 & 0.67 & 14.29 & 0.00 & 0.00 & 4.46 & 2.02 \\
     % & 7.83 & 10.00 & 22.22 & 13.36 & 20.41 & 18.75 & 20.00 & 19.72 & 4.00 & 6.00 & 30.00 & 13.33 & 14.29 & 0.00 & 0.00 & 4.76 & - \\
     HYV & 2.22 & 2.04 & 0.00 & 1.42 & 2.04 & 2.00 & 0.00 & 1.34 & 4.00 & 2.00 & 0.00 & 2.00 & 16.33 & 11.54 & 5.36 & 10.83 & 4.03 \\
     % & 1.96 & 12.00 & 6.67 & 6.88 & 22.45 & 25.00 & 14.00 & 20.48 & 0.00 & 2.00 & 20.00 & 7.33 & 6.12 & 0.00 & 0.00 & 1.91 & - \\
     Wan2.1 & 8.89 & 6.12 & 2.22 & 5.76 & 4.08 & 2.00 & 2.00 & 2.68 & 6.00 & 4.00 & 2.00 & 4.00 & 30.61 & 19.23 & 12.50 & 20.38 & 8.40 \\
     % & 5.88 & 8.00 & 8.89 & 7.59 & 22.45 & 27.08 & 18.00 & 22.51 & 0.00 & 2.00 & 10.00 & 4.00 & 30.61 & 19.23 & 12.50 & 20.38 & - \\
     Wan2.2 & 11.11 & 6.12 & \hlsecond{4.44} & 7.19 & 4.08 & 2.00 & 2.00 & 2.68 & 8.00 & 6.00 & 4.00 & 6.00 & 30.61 & 19.23 & 14.39 & 21.02 & 9.41 \\
     \hline
     \rowcolor{lightgray!20} \multicolumn{18}{c}{{\textit{Commercial Models}}} \\
     Kling 2.1 & 8.89 & 4.08 & 2.22 & 5.04 & 10.20 & 4.00 & 2.00 & 5.37 & 12.00 & 8.00 & 4.00 & 8.00 & 32.65 & 28.85 & 19.64 & 26.75 & 11.60 \\
     Veo 3.1  & \hlsecond{17.78} & \hlsecond{8.16} & \hlsecond{4.44} & \hlsecond{10.07} & \hlsecond{30.61} & \hlsecond{20.00} & \hlsecond{16.00} & \hlsecond{22.15} & \hlfirst{36.00} & \hlsecond{16.00} & \hlsecond{2.00} & \hlsecond{18.00} & \hlfirst{77.55} & \hlfirst{40.38} & \hlfirst{39.29} & \hlfirst{51.59} & \hlsecond{26.05} \\
     Sora 2 & \hlfirst{26.67} & \hlfirst{22.45} & \hlfirst{6.67} & \hlfirst{18.71} & \hlfirst{38.78} & \hlfirst{32.00} & \hlfirst{24.00} & \hlfirst{31.76} & \hlsecond{32.00} & \hlfirst{26.00} & \hlfirst{8.00} & \hlfirst{22.00} & \hlsecond{46.94} & \hlsecond{42.31} & \hlsecond{26.79} & \hlsecond{38.22} & \hlfirst{27.90} \\
     \hlineB{2.5}
   \end{tabular}
  \label{tab:pass@1}
  \vspace{-0.2em}
\end{table*} 
\endgroup

\begingroup
\setlength{\tabcolsep}{1.5pt}
\begin{table*}[t]
% \vspace{-1.4em}
\renewcommand{\arraystretch}{1.5}
  \centering
  \caption{Pass$@5$ performance of open-source models on \ourbench. The \hllfirst{best} and \hllsecond{second best} results are highlighted.}
  \centering
  \vspace{-0.7em}
  \scriptsize
   \begin{tabular}{l cccc cccc cccc cccc c}
     \hlineB{2.5}
     % \rowcolor{CadetBlue!20} 
     \multirow{2}{*}{Model} & \multicolumn{4}{c}{Structural \& Search} & \multicolumn{4}{c}{Spatial \& Visual Pattern} & \multicolumn{4}{c}{Symbolic \& Logical} & \multicolumn{4}{c}{Planning \& Execution} & \multirow{2}{*}{Overall}\\
     % \cline{2-5}\cline{6-10}
     \cmidrule(lr){2-5} \cmidrule(lr){6-9} \cmidrule(lr){10-13} \cmidrule(lr){14-17}
     % \rowcolor{CadetBlue!20} 
     & Easy & Med. & Hard & Over. & Easy & Med. & Hard & Over.  & Easy & Med. & Hard & Over. & Easy & Med. & Hard & Over. & \\
     \hlineB{1.5}
     CVX1.5  & 2.22 & 2.04 & 0.00 & 1.42 & 2.04 & 2.00 & 2.00 & 2.01 & 8.00 & 2.00 & \hllsecond{2.00} & 4.00 & 30.61 & 0.00 & 0.00 & 9.55 & 4.37 \\
     HYV  & \hllsecond{4.44} & 2.04 & 0.00 & 2.16 & \hllsecond{8.16} & \hllsecond{4.00} & \hllsecond{2.00} & 4.70 & 8.00 & \hllsecond{4.00} & \hllsecond{2.00} & 4.67 & 34.69 & 17.31 & 8.93 & 19.75 & 8.07 \\
     Wan2.1  & \hllfirst{24.44} & \hllsecond{16.33} & \hllsecond{8.89} & \hllsecond{16.55} & \hllfirst{14.29} & \hllfirst{6.00} & \hllfirst{4.00} & \hllfirst{8.54} & \hllsecond{10.00} & \hllsecond{4.00} & \hllfirst{4.00} & \hllsecond{6.00} & \hllsecond{44.90} & \hllsecond{26.92} & \hllsecond{19.64} & \hllsecond{29.94} & \hllsecond{15.29} \\
     Wan2.2  & \hllfirst{24.44} & \hllfirst{18.37} & \hllfirst{11.11} & \hllfirst{17.99} & \hllsecond{8.16} & \hllsecond{4.00} & \hllfirst{4.00} & \hllsecond{5.37} & \hllfirst{14.00} & \hllfirst{6.00} & \hllfirst{4.00} & \hllfirst{8.00} & \hllfirst{46.94} & \hllfirst{32.69} & \hllfirst{23.21} & \hllfirst{33.76} & \hllfirst{16.47} \\
     \hlineB{2.5}
   \end{tabular}
  \label{tab:pass@5}
  \vspace{-0.6em}
\end{table*} 
\endgroup

% These results imply that while open-source models are not yet competitive with commercial counterparts, their ability to generate correct solutions under Pass@5 demonstrates untapped reasoning potential. To fully realize this potential, further scaling of training data, model parameters, and reasoning-specific optimizations are needed to improve both the consistency and robustness of their reasoning capabilities.

\subsection{Failure Case Analysis (RQ2)}

\begin{figure*}[!t]
\centering
% \vspace{-0.4em}
\includegraphics[width=\linewidth]{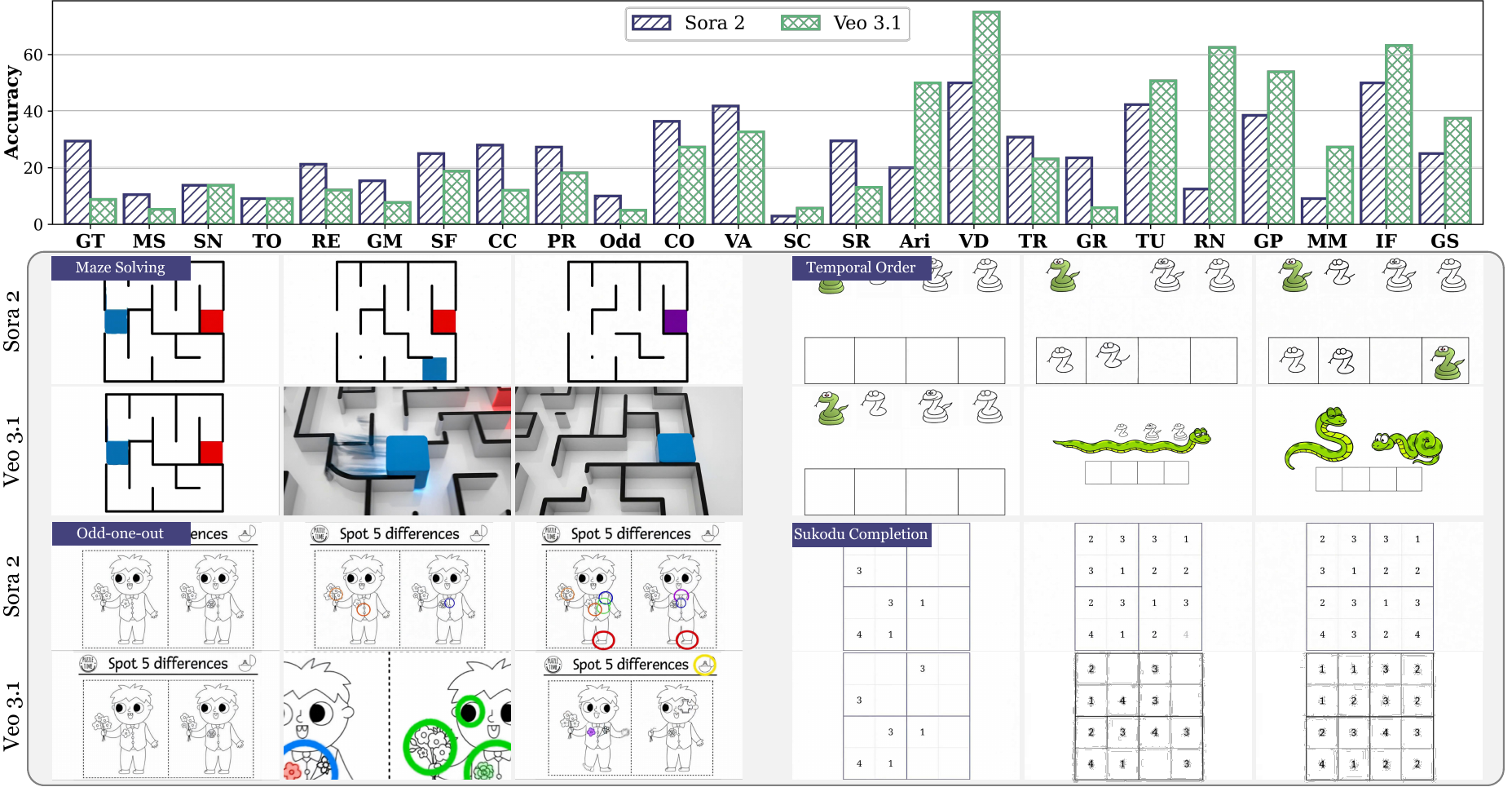}
\vspace{-2em}
\caption{(\textbf{\textit{Top}}) Performance of the best-performing models (\textit{i.e.}, Sora 2 and Veo 3.1) on \ourbench across $24$ tasks. (\textbf{\textit{Bottom}}) Case study of the lowest-performing tasks, \textit{i.e.}, maze solving (MS), temporal ordering (TO), odd-one-out (Odd), and sudoku completion (SC).}
\label{fig:figure6}
\vspace{-0.6em}
\end{figure*}

To answer \textbf{RQ2}, we first conduct an evaluation across $24$ tasks for a more granular analysis, with performance shown in Figure~\ref{fig:figure3}. Subsequently, we further demonstrate failure cases from the tasks with the lowest accuracy, as shown in Figure~\ref{fig:figure6}. We give the following observations:

\vspace{-1em}
\paragraph{Takeaway \ding{184}: Reasoning failures stem from insufficient rule modeling and fine-grained visual feature extraction.} 
% Failure cases in Figure~\ref{fig:figure6} reveal that while Sora 2 and Veo 3.1 excel in general video generation, they struggle with reasoning-specific tasks such as maze solving, temporal ordering, odd-one-out, and sudoku completion. These tasks require explicit logical reasoning, including adherence to scene rules, symbolic manipulation, and subtle categorical reasoning, which current models fail to capture. For instance, in maze solving, generated paths violate spatial boundaries, while in sudoku completion, initial numerical constraints are not preserved.
Figure~\ref{fig:figure6} reveals that while Sora 2 and Veo 3.1 excel in general video generation, they exhibit varying performance across reasoning-specific tasks. For instance, both models achieve relatively high accuracy in tasks, \textit{e.g.}, visual deduction (VD) and instruction following (IF), where reasoning is less dependent on strict rule modeling or symbolic manipulation. However, their performance significantly drops in tasks like maze solving, temporal ordering, odd-one-out, and sudoku completion, which require explicit logical reasoning, including adherence to scene rules, symbolic manipulation, and subtle categorical reasoning. This contrast highlights that current models are better suited for tasks emphasizing general understanding and visual realism, but struggle when reasoning demands structured, rule-based logic.
These failures are likely attributable to two key factors:
(\textbf{\textit{i}}) models struggle to interpret high-level rules, as seen in maze solving tasks where prompts explicitly forbid crossing maze boundaries, yet violations persist; (\textbf{\textit{ii}}) symbolic reasoning requires precise visual feature extraction, but encoders like VAE compress features excessively, losing critical details needed for reasoning.
% (\textbf{\textit{i}}) current video generative models are trained primarily for general generation, with datasets and objectives that prioritize visual realism over logical consistency; (\textbf{\textit{ii}}) existing architectures lack mechanisms to capture fine-grained visual features or enforce structured reasoning, making them ill-suited for symbolic or logic-intensive tasks. 
Addressing these gaps will require explicit task rule encoding, reinforcement learning for process-level optimization, and more fine-grained visual feature representations and structured processing.

\begingroup
\setlength{\tabcolsep}{1.8pt}
\begin{table*}[!t]
\vspace{-0.4em}
\renewcommand{\arraystretch}{1.5}
  \centering
  \caption{Evaluation on \ourbench with \ourmethod. We \textbf{bold} the best results. Qualitative results are in \textbf{Appendix}~\S\ref{app:exhibition}.}
  \centering
  \vspace{-0.7em}
  \scriptsize
   \begin{tabular}{l cccc cccc cccc cccc c}
     \hlineB{2.5}
     % \rowcolor{CadetBlue!20} 
     \multirow{2}{*}{Model} & \multicolumn{4}{c}{Structural \& Search} & \multicolumn{4}{c}{Spatial \& Visual Pattern} & \multicolumn{4}{c}{Symbolic \& Logical} & \multicolumn{4}{c}{Planning \& Execution} & \multirow{2}{*}{Overall}\\
     % \cline{2-5}\cline{6-10}
     \cmidrule(lr){2-5} \cmidrule(lr){6-9} \cmidrule(lr){10-13} \cmidrule(lr){14-17}
     % \rowcolor{CadetBlue!20} 
     & Easy & Med. & Hard & Over. & Easy & Med. & Hard & Over.  & Easy & Med. & Hard & Over. & Easy & Med. & Hard & Over. & \\
     \hlineB{1.5}
     % CogVideoX1.5 \citep{yang2024cogvideox} & - & - & - & - & - & - & - & - & - & - & - & - & - & - & - & - & - \\
     % Wan2.2 \citep{wan2025} & - & - & - & - & - & - & - & - & - & - & - & - & - & - & - & - & - \\
     % \hline
     HunyuanVideo & 2.22 & 2.04 & 0.00 & 1.42 & 2.04 & 2.00 & 0.00 & 1.34 & 4.00 & 2.00 & 0.00 & 2.00 & 16.33 & 11.54 & 5.36 & 10.83 & 4.03 \\
     $+$ Pre-Rewriter & 4.44 & 2.04 & 0.00 & 2.16 & 6.12 & 0.00 & 0.00 & 2.01 & 6.00 & 4.00 & 0.00 & 3.33 & 20.41 & 11.54 & 1.79 & 10.83 & 4.71 \\
     $+$ Post-Rewriter  & 8.89 & 4.08 & 0.00 & 4.32 & 6.12 & 4.00 & 2.00 & 4.03 & 8.00 & 6.00 & 0.00 & 4.67 & 20.41 & 13.46 & 5.36 & 12.74 & 6.55 \\
     \rowcolor{cyan!10}
     $+$ \textbf{\ourmethod (Ours)} & \textbf{13.33} & \textbf{6.12} & \textbf{4.44} & \textbf{7.91} & \textbf{8.16} & \textbf{6.00} & \textbf{2.00} & \textbf{5.37} & \textbf{12.00} & \textbf{6.00} & \textbf{2.00} & \textbf{6.67} & \textbf{36.73} & \textbf{21.15} & \textbf{12.50} & \textbf{22.93} & \textbf{10.25} \\
     \hline
     Wan2.1 & 8.89 & 6.12 & 2.22 & 5.76 & 4.08 & 2.00 & 2.00 & 2.68 & 6.00 & 4.00 & 2.00 & 4.00 & 30.61 & 19.23 & 12.50 & 20.38 & 8.40 \\
     $+$ Pre-Rewriter  & 11.11 & 8.16 & 2.22 & 7.19 & 8.16 & 4.00 & 4.00 & 5.37 & 10.00 & 2.00 & 2.00 & 4.00 & 38.78 & 25.00 & 14.29 & 25.48 & 10.76 \\
     $+$ Post-Rewriter & 15.56 & 8.16 & 4.44 & 9.35 & 12.24 & 6.00 & 4.00 & 7.38 & 8.00 & 4.00 & 2.00 & 4.67 & 36.73 & 26.92 & 16.07 & 26.11 & 12.10 \\
     \rowcolor{cyan!10}
     $+$ \textbf{\ourmethod (Ours)} & \textbf{28.89} & \textbf{20.41} & \textbf{8.89} & \textbf{19.42} & \textbf{16.33} & \textbf{8.00} & \textbf{6.00} & \textbf{10.07} & \textbf{14.00} & \textbf{8.00} & \textbf{4.00} & \textbf{8.67} & \textbf{48.98} & \textbf{30.77} & \textbf{23.21} & \textbf{33.76} & \textbf{18.15} \\
     \hlineB{2.5}
   \end{tabular}
  \label{tab:videotpo}
  % \vspace{-1em}
\end{table*} 
\endgroup

\subsection{Results with \textbf{\ourmethod} (RQ3)}

Building on the above observations, we further sought to investigate whether test-time scaling could deliver more efficient inference optimization than large-scale training. To answer \textbf{RQ3}, we conducted a comprehensive evaluation of our proposed \ourmethod in Table~\ref{tab:videotpo}, alongside two baseline strategies: pre-rewriter based on \citep{vertex-ai-video-gen-prompt-guide}, and post-rewriter based on \citep{madaan2023self}. The following observation is drawn:

\vspace{-1em}
\paragraph{Takeaway \ding{185}: \ourmethod is an effective test-time video generation reasoning enhancer.}
Table~\ref{tab:videotpo} demonstrates that our \ourmethod consistently improves reasoning accuracy across all dimensions and difficulty levels, outperforming both the base models and baseline strategies. For instance, applying \ourmethod to HunyuanVideo improves overall performance from $4.03\%$ to $10.25\%$, while for Wan2.1, the improvement is even more pronounced, increasing from $8.40\%$ to $18.15\%$. These gains highlight the ability of \ourmethod to refine inference-time reasoning without requiring additional training.
Furthermore, pre-rewriter and post-rewriter strategies also yield performance improvements. This indicates that test-time scaling can effectively unlock reasoning capabilities for video generation.
% The findings strongly validate the effectiveness of \ourmethod as a practical and efficient test-time reasoning enhancer, setting a new benchmark for inference-time optimization in video generation tasks.

\subsection{Further Analysis}

\begin{figure*}[!t]
\centering
% \vspace{-0.4em}
\includegraphics[width=\linewidth]{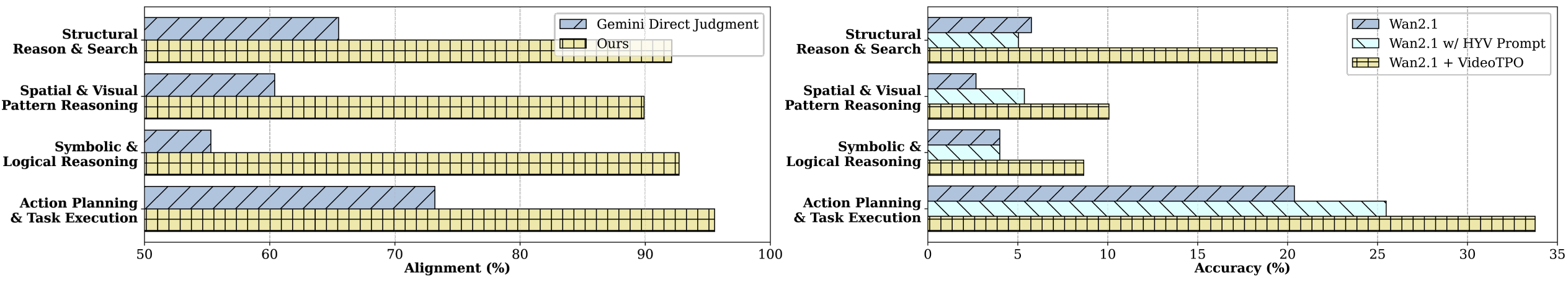}
\vspace{-2em}
\caption{(\textbf{\textit{Left}}) Agreement between our metrics and human judgments in Wan 2.1 evaluation; (\textbf{\textit{Right}}) Comparison of different prompt strategies, \texttt{\small`w/ HYV Prompt'} indicates using \ourmethod optimized prompts based on HunyuanVideo. }
\label{fig:figure7}
\vspace{-0.6em}
\end{figure*}

\paragraph{Analysis of Metric-Human Alignment.} To evaluate the reliability of our proposed metrics, we compare the alignment with human judgments, as shown in Figure~\ref{fig:figure7} (\textit{Left}). Our metrics demonstrate high alignment with human assessments, validating the robustness of our metrics in capturing reasoning-specific task performance, offering a reliable alternative to manual evaluation.

\vspace{-1em}
\paragraph{Analysis of \ourmethod Refined Prompt.} We further evaluate the impact of prompt optimization by comparing Wan2.1 with two refined prompts: \texttt{\small`w/ HYV Prompt'} (optimized by \ourmethod on HunyuanVideo) and \texttt{\small`+ \ourmethod'}. As shown in Figure~\ref{fig:figure7} (\textit{Right}), \texttt{\small`Wan2.1 w/ HYV Prompt'} shows limited improvement or even degradation, while \texttt{\small`Wan2.1 + \ourmethod'} achieves significant gains across all dimensions. Beyond validating the effectiveness of our \ourmethod, this further demonstrates that different models exhibit varying preferences for prompts.
% and the prompts carefully designed for \ourbench ensure compatibility with most models.

\section{Conclusion}

In this work, we present \textbf{\ourbench}, a hierarchical benchmark designed to evaluate reasoning capabilities of I2V generation models across four dimensions: \textit{Structural Reasoning \& Search}, \textit{Spatial \& Visual Pattern Reasoning}, \textit{Symbolic \& Logical Reasoning}, and \textit{Action Planning \& Task Execution}. With $595$ samples across $24$ task scenarios and $3$ difficulty levels, \ourbench provides a comprehensive suite for benchmarking the reasoning capabilities in video generation models. Our evaluation reveals that while commercial models demonstrate stronger and more consistent reasoning capabilities, open-source models show promising yet unstable performance. To this end, we propose \textbf{\ourmethod}, a lightweight test-time strategy leveraging multi-pass generation and preference alignment to unlock model potential without additional training, achieving fine-grained optimization at test time.

% \clearpage
\newpage
\bibliographystyle{assets/plainnat}
\bibliography{paper}

\clearpage
\newpage
\beginappendix

\appendix
\crefalias{section}{appendix}

\section{More Details of Evaluation Dimension}
\label{app:dimension}

\subsection{Motivation for Each Scenario}

To comprehensively evaluate visual reasoning in video generative models, we design $24$ diverse task scenarios across four key dimensions. Each scenario is carefully crafted to challenge specific aspects of visual reasoning, ensuring a systematic assessment of models’ ability to perform beyond general video generation. Below, we outline the motivation for each scenario:

\vspace{-1em}
\paragraph{Structural Reasoning \& Search.} Structural reasoning tasks assess models' ability to understand and navigate complex spatial structures, sequences, and rules, which are critical for reasoning in dynamic environments.

\vspace{-0.4em}
\begin{tcolorbox}[breakable, colback=gray!4, colframe=black,  width=1\textwidth]
\vspace{-1.5mm}
\begin{itemize}[leftmargin=0.8mm]
    \item {\small\textit{Graph Traversal}: Tests the model's capability to explore structured graphs and identify valid traversal paths, simulating real-world spatial reasoning.}
    \item {\small\textit{Maze Solving}: Challenges models to navigate through constrained environments, requiring spatial planning and decision-making.}
    \item {\small\textit{Sorting Numbers}: Evaluates logical ordering of visual elements, emphasizing reasoning over numerical structures in dynamic contexts.}
    \item {\small\textit{Temporal Ordering}: Assesses the model's ability to infer sequential relationships between events or frames.}
    \item {\small\textit{Rule Extrapolation}: Tests the model's understanding of abstract rules and its ability to generalize them to new scenarios.}
    \item {\small\textit{Game Move}: Simulates decision-making in strategic games, requiring models to predict valid moves based on spatial and logical reasoning.}
\end{itemize}
% \vspace{-2mm}
\end{tcolorbox}

\vspace{-0.6em}
\paragraph{Spatial \& Visual Pattern Reasoning.} These scenarios focus on recognizing patterns, relationships, and visual consistencies, which are foundational to reasoning in visual contexts.

\vspace{-0.4em}
\begin{tcolorbox}[breakable, colback=gray!4, colframe=black,  width=1\textwidth]
\vspace{-1.5mm}
\begin{itemize}[leftmargin=0.8mm]
    \item {\small\textit{Shape Fitting}: Challenges models to match shapes into predefined spaces, testing spatial alignment and pattern recognition.}
    \item {\small\textit{Connecting Colors}: Evaluates the ability to identify and connect visually related elements based on color patterns.}
    \item {\small\textit{Pattern Recognition}: Assesses model's capacity to detect recurring patterns and infer underlying rules.}
    \item {\small\textit{Odd-one-out}: Tests model's ability to identify anomalies in visual sets, requiring attention to detail and comparative reasoning.}
    \item {\small\textit{Counting Objects}: Focuses on numerical reasoning by evaluating the model's ability to count and quantify visual elements.}
    \item {\small\textit{Visual Analogy}: Assesses abstract reasoning by requiring models to identify analogical relationships between visual objects.}
\end{itemize}
% \vspace{-2mm}
\end{tcolorbox}

\vspace{-0.6em}
\paragraph{Symbolic \& Logical Reasoning.} Symbolic reasoning tasks test the ability to understand abstract symbols, logical rules, and numerical relationships.

\vspace{-0.4em}
\begin{tcolorbox}[breakable, colback=gray!4, colframe=black,  width=1\textwidth]
\vspace{-1.5mm}
\begin{itemize}[leftmargin=0.8mm]
    \item {\small\textit{Sudoku Completion}: Challenges models to complete structured puzzles based on logical constraints, testing symbolic reasoning.}
    \item {\small\textit{Symbolic Reasoning}: Evaluates the model's ability to infer relationships between abstract symbols and make logical deductions.}
    \item {\small\textit{Arithmetic}: Tests numerical reasoning by requiring models to solve basic arithmetic problems presented visually.}
    \item {\small\textit{Visual Deduction}: Assesses the ability to infer logical conclusions from visual cues, such as completing partially visible objects.}
    \item {\small\textit{Transitive Reasoning}: Challenges models to infer indirect relationships between elements based on transitive logic.}
    \item {\small\textit{Game Rule}: Evaluates understanding of abstract rules and their application in visual environments.}
\end{itemize}
% \vspace{-2mm}
\end{tcolorbox}

\vspace{-0.6em}
\paragraph{Action Planning \& Task Execution.} These tasks simulate real-world scenarios requiring multi-step planning, execution, and adaptability in dynamic environments.

\vspace{-0.4em}
\begin{tcolorbox}[breakable, colback=gray!4, colframe=black,  width=1\textwidth]
\vspace{-1.5mm}
\begin{itemize}[leftmargin=0.8mm]
    \item {\small\textit{Tool Use}: Assesses models' ability to infer the correct use of tools based on visual cues and task requirements.}
    \item {\small\textit{Robot Navigation}: Challenges models to plan and execute navigation in complex spatial environments, simulating robotic reasoning.}
    \item {\small\textit{Goal-directed Planning}: Tests multi-step planning towards achieving specific goals in dynamic settings.}
    \item {\small\textit{Multi-step Manipulation}: Evaluates the ability to coordinate and execute sequential actions to manipulate objects.}
    \item {\small\textit{Instruction Following}: Assesses models' capacity to interpret visual instructions and execute tasks accordingly.}
    \item {\small\textit{Game Strategy}: Challenges strategic reasoning by requiring models to plan and execute moves in visually dynamic games.}
\end{itemize}
% \vspace{-2mm}
\end{tcolorbox}

\begin{figure*}[!t]
\centering
\vspace{-0.4em}
\includegraphics[width=\linewidth]{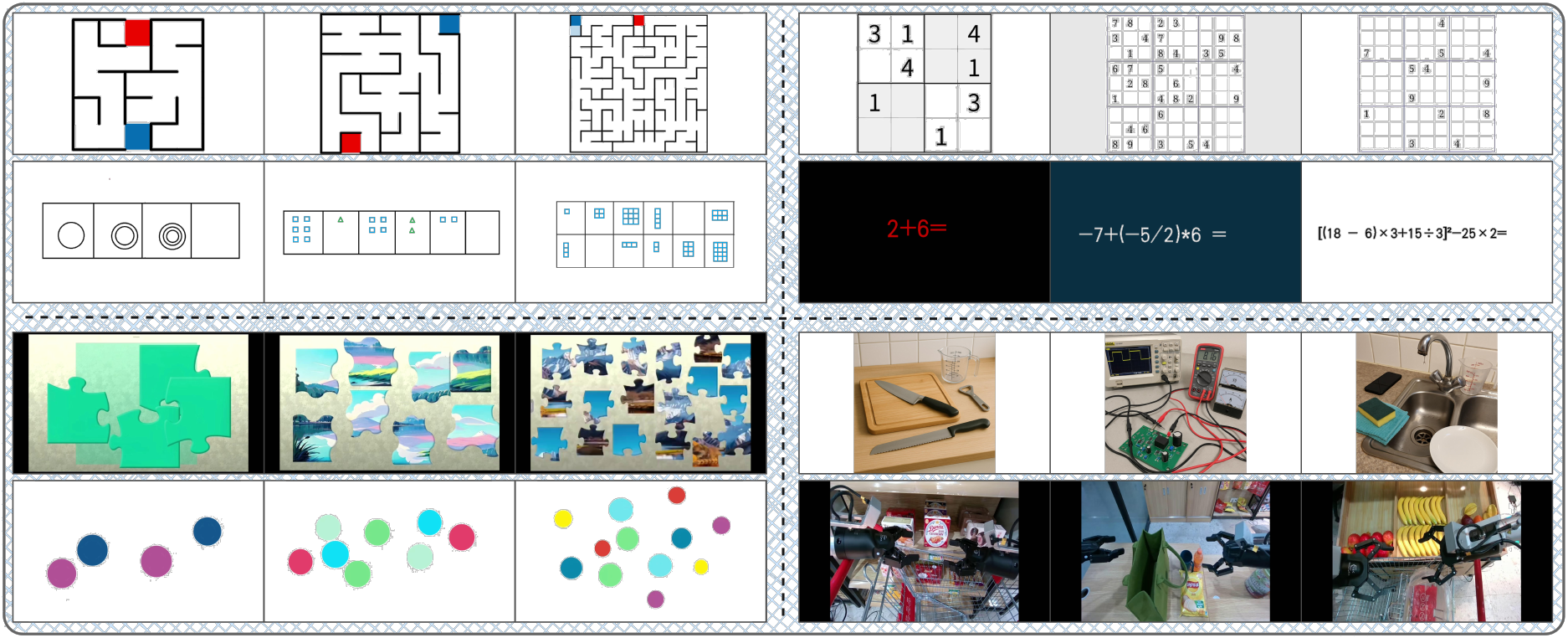}
\vspace{-1.9em}
 \caption{Data demonstration across easy, medium, and hard. (\textbf{\textit{Top Left}}) Structural Reasoning \& Search. (\textbf{\textit{Top Right}}) Symbolic \& Logical Reasoning. (\textbf{\textit{Bottom Left}}) Spatial \& Visual Pattern Reasoning. (\textbf{\textit{Bottom Right}}) Action Planning \& Task Execution. }
\label{fig:app_a_1}
% \vspace{-0.5em}
\end{figure*}

\subsection{Data Demonstration}

Here, we present examples of our \ourbench in Figure~\ref{fig:app_a_1} to provide a more vivid illustration of the three difficulty levels: easy, medium, and hard.

\section{More Details of Prompt Suite}
\label{app:prompt}

% \vspace{-1em}
\subsection{Prompt Generation}

We adopt Gemini-2.5-Pro \citep{deepmind_gemini_pro_2024} as a powerful assistant to generate an initial version of the inference prompt for our \ourbench. Here we further detail the task prompt for each dimension:

\begin{tcolorbox}[notitle, sharp corners, breakable, colframe=MidnightBlue!80, colback=gray!10, 
       boxrule=3pt, boxsep=0.5pt, enhanced, 
       shadow={3pt}{-3pt}{0pt}{opacity=1,newgray},
       title={Structural Reasoning \& Search}]\label{box:prompt1}
       \footnotesize
       \setstretch{1}
       {\fontfamily{pcr}\selectfont
\begin{lstlisting}
"""You are a senior researcher in computer vision. You are tasked with generating detailed prompts for Image-to-Video (I2V) data samples that evaluate Structural Reasoning & Search abilities.
You are given two images: {initial_image} shows the initial state, and {target_image} shows the target state. The corresponding task is {task}.
Generate a detailed, narratively rich prompt describing how the main subject logically evolves from the initial to the target state.  

Key points to emphasize:  
- Center on video content, avoiding overly directive instructions.
- Clearly define the start and end states without revealing the exact solution path, maintaining goal clarity.  
- Imply hidden constraints or rules that the model must infer to understand the transformation.  
- Ensure the prompt describes a task that unfolds logically and coherently over time, highlighting temporal progression.  
- Keep the prompt length under 150 tokens.

Describe the transformation as a logical exploration or structured problem-solving journey, inviting the model to infer intermediate steps and rules that connect the two states.
"""

\end{lstlisting}
}
\end{tcolorbox}

\begin{tcolorbox}[notitle, sharp corners, breakable, colframe=MidnightBlue!80, colback=gray!10, 
       boxrule=3pt, boxsep=0.5pt, enhanced, 
       shadow={3pt}{-3pt}{0pt}{opacity=1,newgray},
       title={Spatial \& Visual Pattern Reasoning}]\label{box:prompt2}
       \footnotesize
       \setstretch{1}
       {\fontfamily{pcr}\selectfont
\begin{lstlisting}
"""You are a senior researcher in computer vision. You are tasked with generating detailed prompts for Image-to-Video (I2V) data samples that evaluate Spatial & Visual Pattern Reasoning abilities. You are given two images: {initial_image} shows the initial spatial arrangement, and {target_image} shows the target arrangement. The corresponding task is {task}. 
Generate a vivid, descriptive prompt explaining how the main subject spatially transforms from the initial to the target state.  

Key points to emphasize:  
- Center on video content, avoiding overly directive instructions.
- Provide rich visual descriptions of shapes, colors, positions, and spatial relationships to enhance visual specificity.  
- Encourage recognition and extension of visual patterns, such as shape fitting, rotations, or color connections.  
- Allow for open-ended interpretations or multiple valid transformations, without restricting to a single solution.  
- Keep the prompt length under 150 tokens.

Narrate the spatial evolution as a dynamic visual story, focusing on how the subject's spatial configuration changes over time.
"""

\end{lstlisting}
}
\end{tcolorbox}

\begin{tcolorbox}[notitle, sharp corners, breakable, colframe=MidnightBlue!80, colback=gray!10, 
       boxrule=3pt, boxsep=0.5pt, enhanced, 
       shadow={3pt}{-3pt}{0pt}{opacity=1,newgray},
       title={Symbolic \& Logical Reasoning}]\label{box:prompt3}
       \footnotesize
       \setstretch{1}
       {\fontfamily{pcr}\selectfont
\begin{lstlisting}
"""You are a senior researcher in computer vision. You are tasked with generating detailed prompts for Image-to-Video (I2V) data samples that evaluate Symbolic & Logical Reasoning abilities. You are given two images: {initial_image} shows the initial symbolic or logical state, and {target_image} shows the target state. The corresponding task is {task}.
Generate a detailed, narratively engaging prompt describing how the symbolic elements or logical conditions in the initial image evolve into those in the target image.  

Key points to emphasize:
- Center on video content, avoiding overly directive instructions.
- Avoid explicitly stating the rules; instead, imply constraints so that the model discovers them implicitly.  
- Integrate symbolic reasoning tightly with the visual elements present in the images.  
- Ensure the task involves a clear logical progression or sequence of reasoning steps connecting the two states.  
- Keep the prompt length under 150 tokens.

Describe the transformation as a story of abstract reasoning and symbolic manipulation unfolding through logical inference.
"""

\end{lstlisting}
}
\end{tcolorbox}

\begin{tcolorbox}[notitle, sharp corners, breakable, colframe=MidnightBlue!80, colback=gray!10, 
       boxrule=3pt, boxsep=0.5pt, enhanced, 
       shadow={3pt}{-3pt}{0pt}{opacity=1,newgray},
       title={Action Planning \& Task Execution}]\label{box:prompt4}
       \footnotesize
       \setstretch{1}
       {\fontfamily{pcr}\selectfont
\begin{lstlisting}
"""You are a senior researcher in computer vision. You are tasked with generating detailed prompts for Image-to-Video (I2V) data samples that evaluate Action Planning & Task Execution abilities. You are given two images: {initial_image} shows the initial scenario, and {target_image} shows the final scenario. The corresponding task is {task}.
Generate a richly descriptive, narrative prompt explaining how the main subject plans and executes a sequence of actions to reach the target state.  

Key points to emphasize:
- Center on video content, avoiding overly directive instructions.
- Define the overall goal clearly while leaving intermediate steps implicit, encouraging goal-oriented interpretation.  
- Highlight the necessity of multi-step reasoning and sequential action planning.  
- Emphasize causal relationships and logical cause-and-effect connections between actions and outcomes.  
- Keep the prompt length under 150 tokens.

Frame the transformation as a purposeful, temporally coherent journey of task execution and goal fulfillment.
"""

\end{lstlisting}
}
\end{tcolorbox}

\subsection{Case Study}

\begin{figure*}[!b]
\centering
\vspace{-0.4em}
\includegraphics[width=\linewidth]{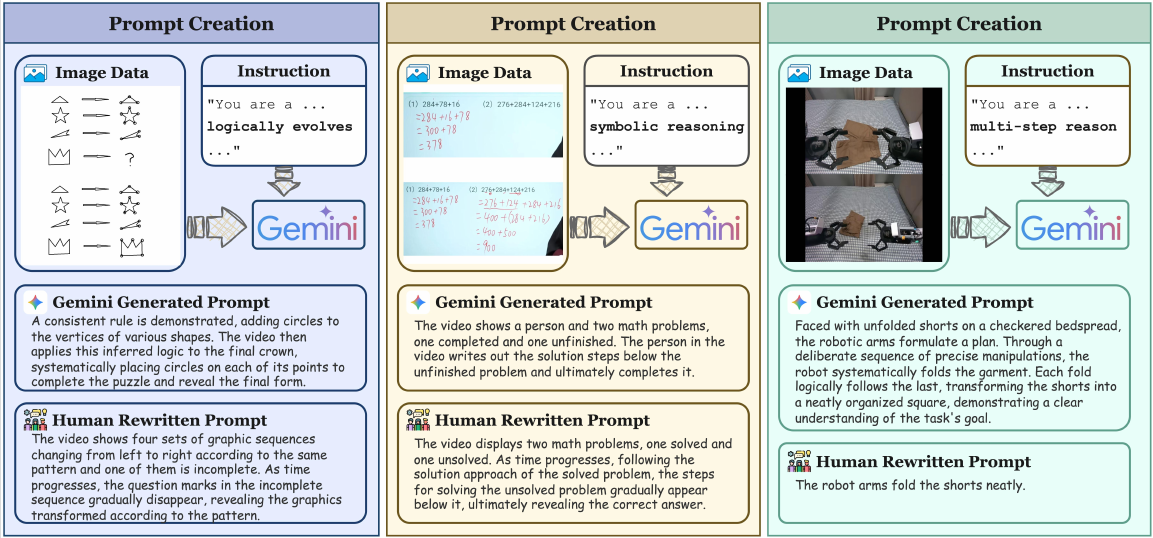}
\vspace{-1.9em}
 \caption{Example demonstrations of our prompt creation process. }
\label{fig:app_b_2}
\vspace{-0.4em}
\end{figure*}

Here we provide case studies on our prompt creation process in Figure~\ref{fig:app_b_2}.

\section{More Details of Metric Suite}
\label{app:metric}

Unlike the metrics commonly used for evaluating video generation models (\textit{e.g.}, temporal coherence, semantic alignment \citep{huang2024vbench++, huang2024vbench}), most visual reasoning tasks often have verifiable targets. However, unlike LLM reasoning, which can rely on expert models (\textit{e.g.}, GPT-4o \citep{wang2024measuring, lu2024mathvista}) at the text level, evaluating visual reasoning requires models to demonstrate a wide range of visual capabilities, \textit{e.g.}, OCR, counting, and tracking. This makes it challenging to achieve a comprehensive evaluation using a single expert model. To this end, we design task-specific metrics to accurately and systematically assess different types of tasks. 
% For technical details, please refer to our code.

\subsection{Final-State Validation} 
\paragraph{OpenCV-based Metrics.} To evaluate visual reasoning tasks with clear and verifiable targets, we leverage OpenCV-based \citep{opencv} metrics tailored to specific task types. These metrics are designed to assess the model's ability to perform nuanced visual operations such as edge detection, contour extraction, object segmentation, and OCR.
\begin{itemize}[leftmargin=1.3em]
    \item[\ding{182}] \textit{Sudoku Recognition}: This metric evaluates the ability to extract and interpret the digits within a Sudoku grid from an image or video frame. The process involves detecting the grid structure via edge detection and contour approximation, applying perspective transformation, and segmenting the grid into cells. The extracted digit matrix is compared against the ground truth for correctness.
    \begin{figure}[!h]
    \centering
    \vspace{-0.8em}
    \hspace*{0.42cm}
    \includegraphics[width=0.54\linewidth]{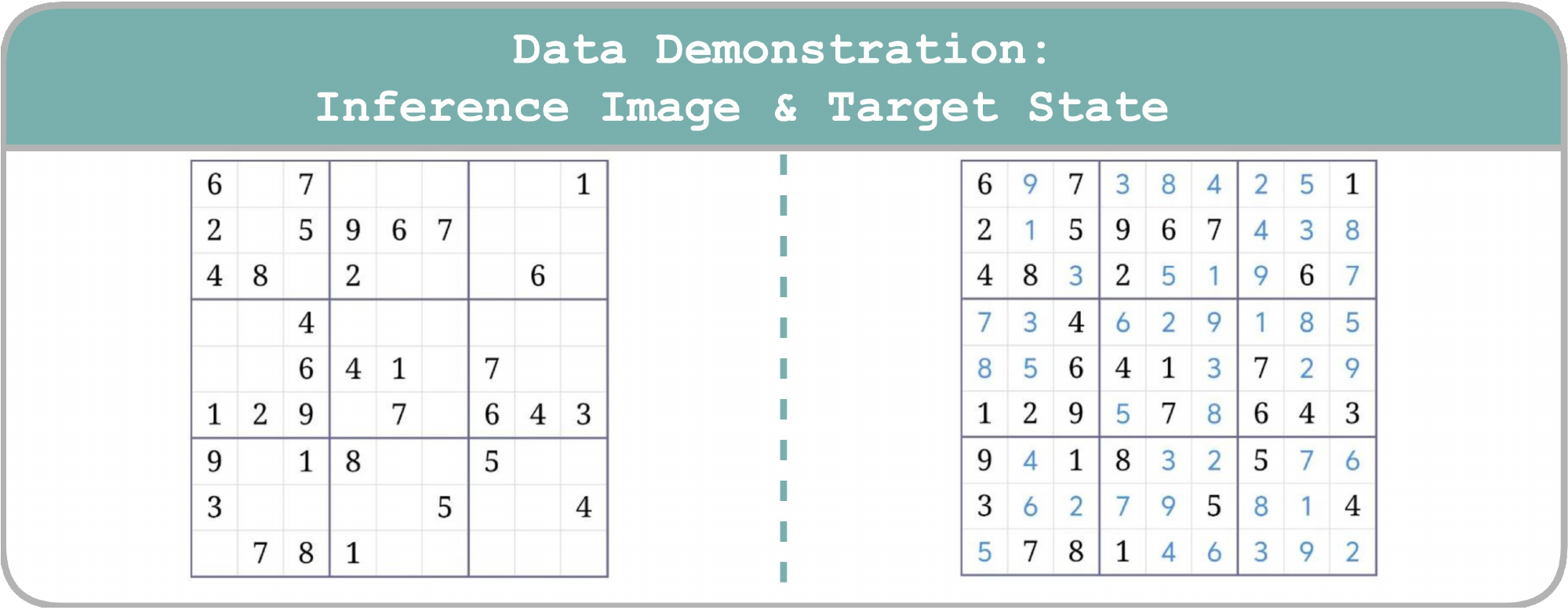}
    \vspace{-1.2em}
    \end{figure}
    \item[\ding{183}] \textit{Mathematical Evaluation}: For tasks involving mathematical equations, this metric assesses the accuracy of OCR-based text recognition and the semantic equivalence of mathematical expressions. After preprocessing the image (\textit{e.g.}, binarization), the recognized text is parsed and evaluated. The comparison accounts for both exact textual matches and equivalence in computed results, ensuring a comprehensive assessment of the model's reasoning capabilities.
    \begin{figure}[!h]
    \centering
    \vspace{-0.8em}
    \hspace*{0.42cm}
    \includegraphics[width=0.54\linewidth]{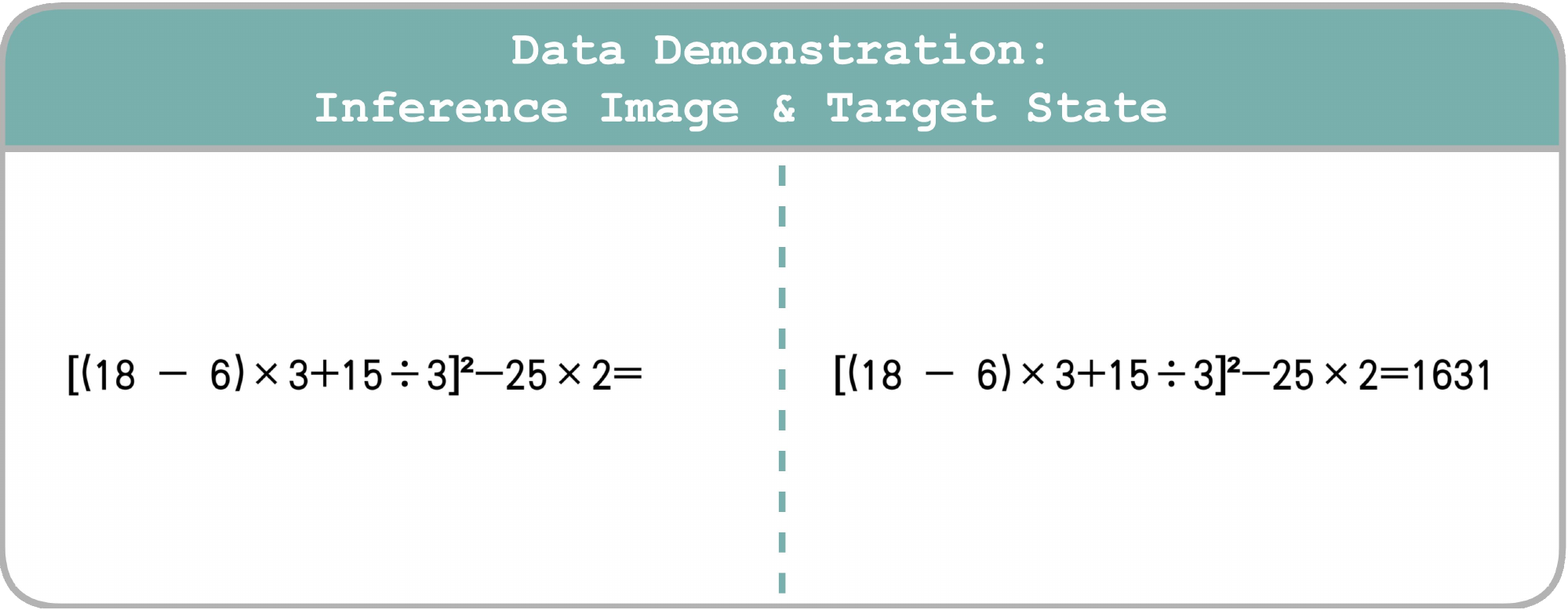}
    \vspace{-1.2em}
    \end{figure}
    \item[\ding{184}] \textit{Visual Multiple Choice}: This metric is designed for tasks requiring the identification of correct answers from visual cues, such as detecting red boxes containing letters. It utilizes color segmentation in HSV space to identify candidate regions and applies OCR to extract the letter within each detected box. The correctness is determined by matching the extracted letter with the ground truth answer.
    \begin{figure}[!h]
    \centering
    \vspace{-0.8em}
    \hspace*{0.42cm}
    \includegraphics[width=0.54\linewidth]{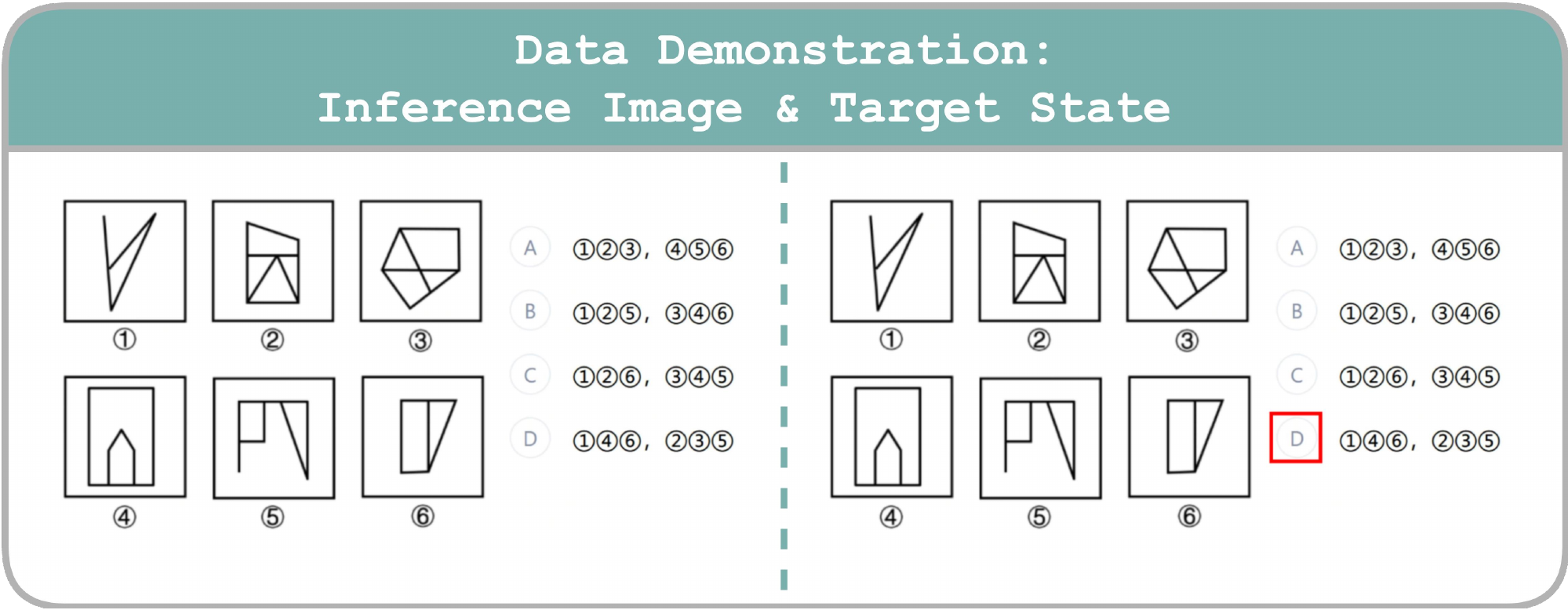}
    \vspace{-1.2em}
    \end{figure}
    \item[\ding{185}] \textit{Numeric Sequence Completion}: For tasks requiring the completion of numerical sequences, this metric evaluates the accuracy of OCR-based recognition of digits. Through preprocessing and binarization, the sequence is extracted from the video frame and compared with the ground truth. This metric focuses on precise textual recognition and sequence matching.
    \begin{figure}[!h]
    \centering
    \vspace{-0.8em}
    \hspace*{0.42cm}
    \includegraphics[width=0.54\linewidth]{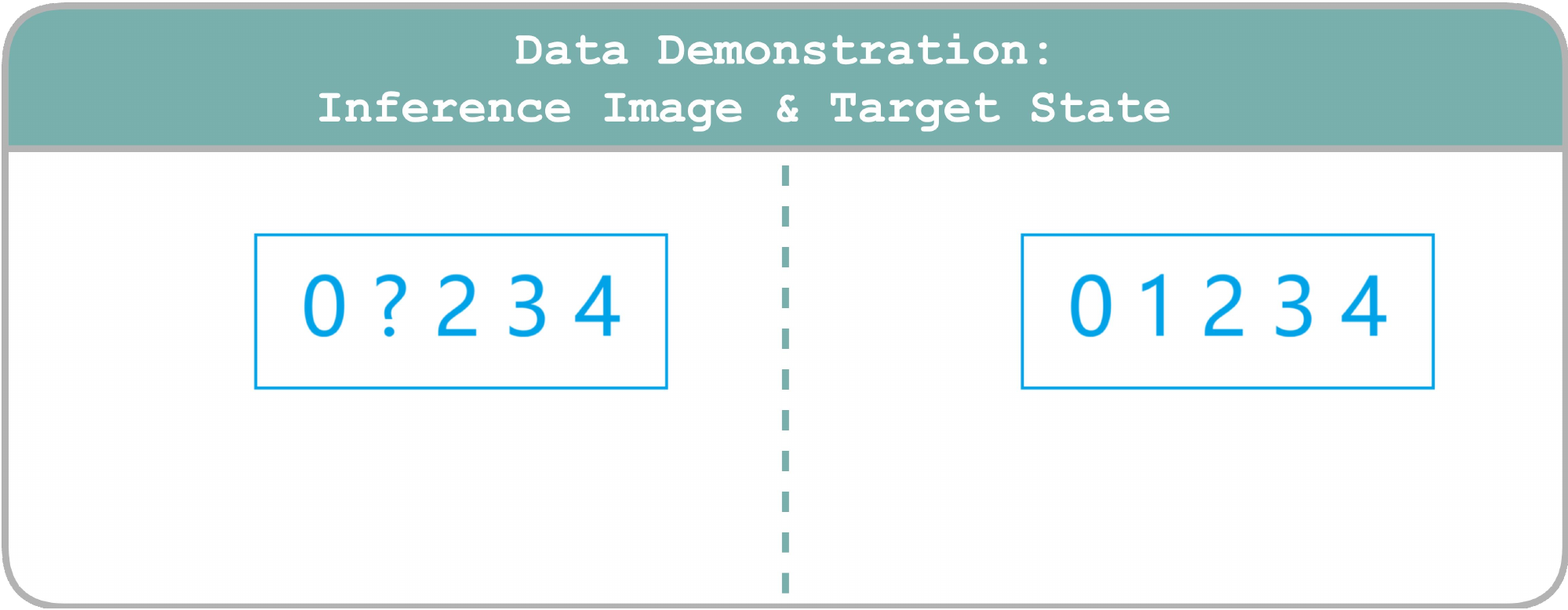}
    \vspace{-1.2em}
    \end{figure}
    \item[\ding{186}] \textit{Graphic Sorting Tasks}: This metric assesses the model's ability to detect and compare graphical elements, such as blue bars in sorting tasks. Using color segmentation and contour analysis, the heights of bars are measured and compared against the ground truth. The evaluation accounts for both the number of detected bars and their relative heights, ensuring alignment with the expected order.
    \begin{figure}[!h]
    \centering
    \vspace{-0.8em}
    \hspace*{0.42cm}
    \includegraphics[width=0.54\linewidth]{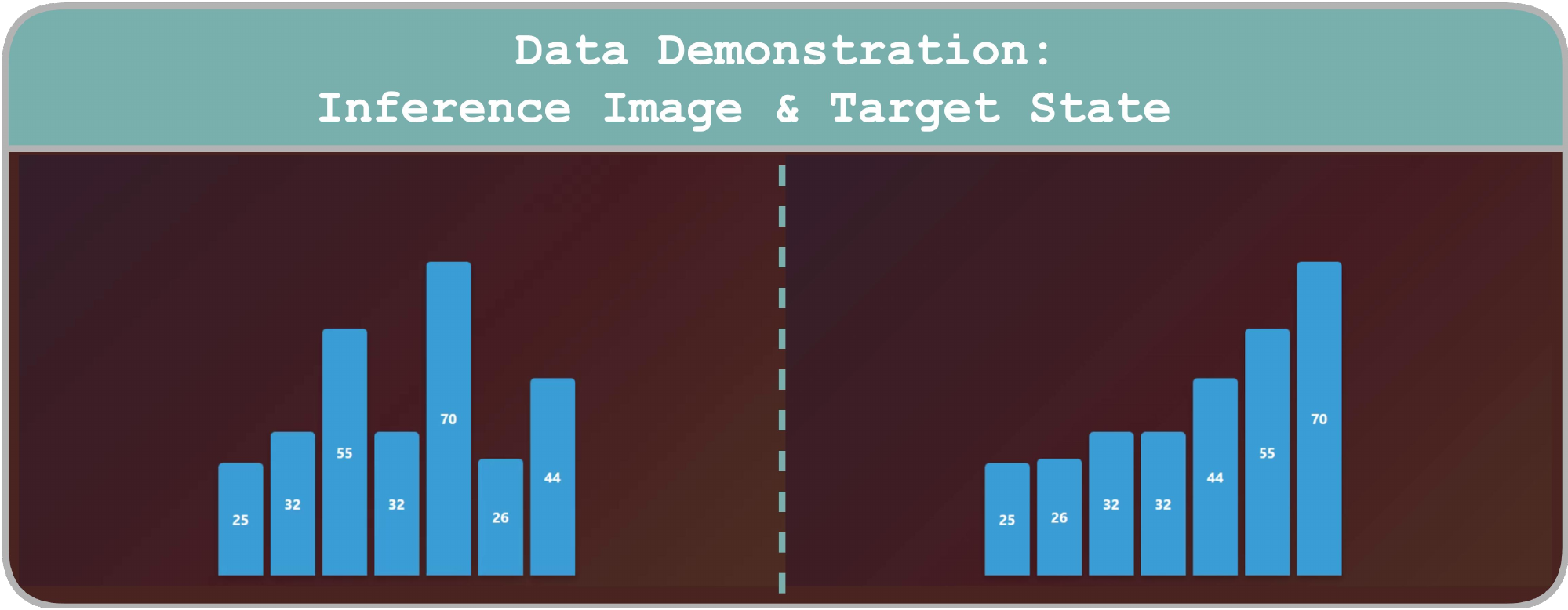}
    \vspace{-1.2em}
    \end{figure}
    \item[\ding{187}] \textit{Match-3-like Games}: For visual tasks resembling games (\textit{e.g.}, "match-3" or elimination games), this metric compares the structural and pixel-level similarity between the final frame and the ground truth. Edge detection and SSIM are used to evaluate the overlap in patterns and overall image alignment, ensuring the model's output adheres to the expected configuration.
    \begin{figure}[!h]
    \centering
    \vspace{-0.8em}
    \hspace*{0.42cm}
    \includegraphics[width=0.54\linewidth]{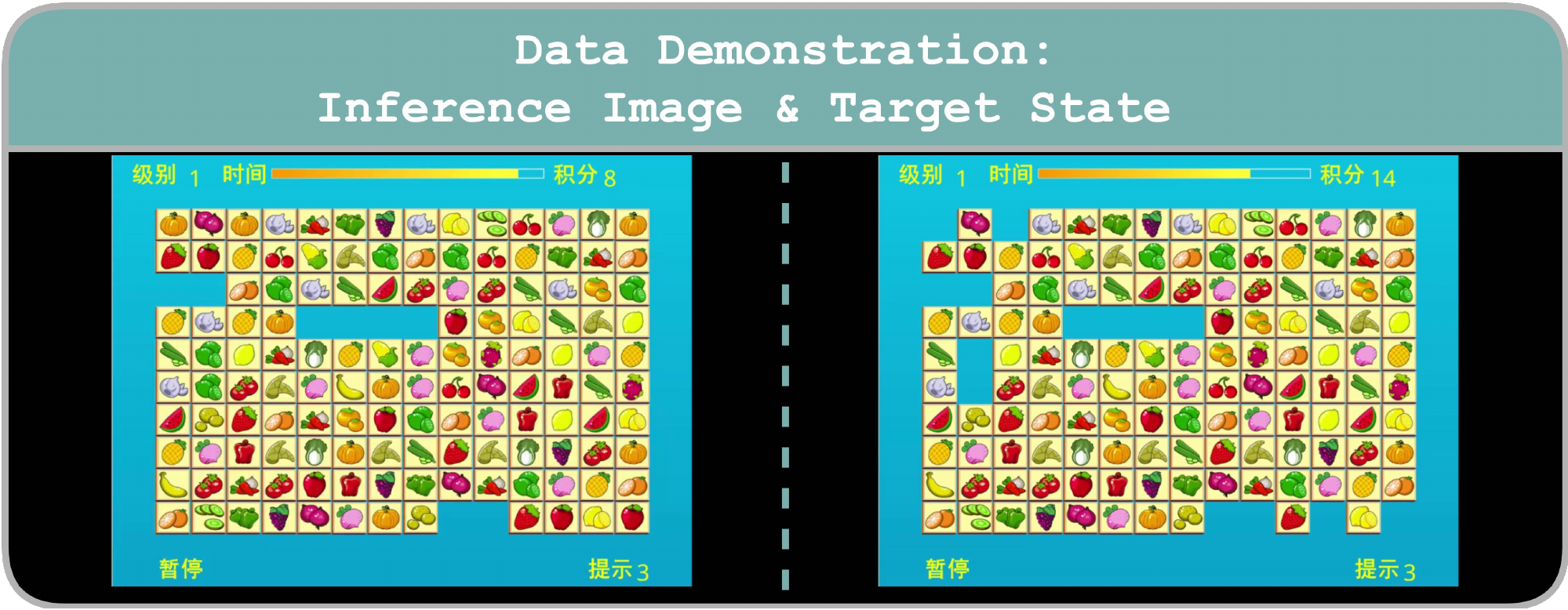}
    \vspace{-0.6em}
    \end{figure}
\end{itemize}

\vspace{-1em}
\paragraph{DINO-based Metrics.} To evaluate tasks requiring complex visual reasoning and spatial understanding, we design a set of metrics based on DINO \citep{caron2021emerging}. These metrics are particularly suited for tasks that involve structured visual patterns, such as \textit{completing shape sequences}, \textit{refining sketches}, \textit{organizing temporal events}, \textit{solving puzzles}, \textit{spatial reasoning} (\textit{e.g.}, mirroring, rotation), and \textit{board game recognition}. By leveraging DINO's ability to extract robust and high-level semantic features, we ensure that the evaluation is both adaptable and precise.

\begin{wrapfigure}{r}{0.48\textwidth}
\vspace{-1.2em}
 \centering
 \includegraphics[width=\linewidth]{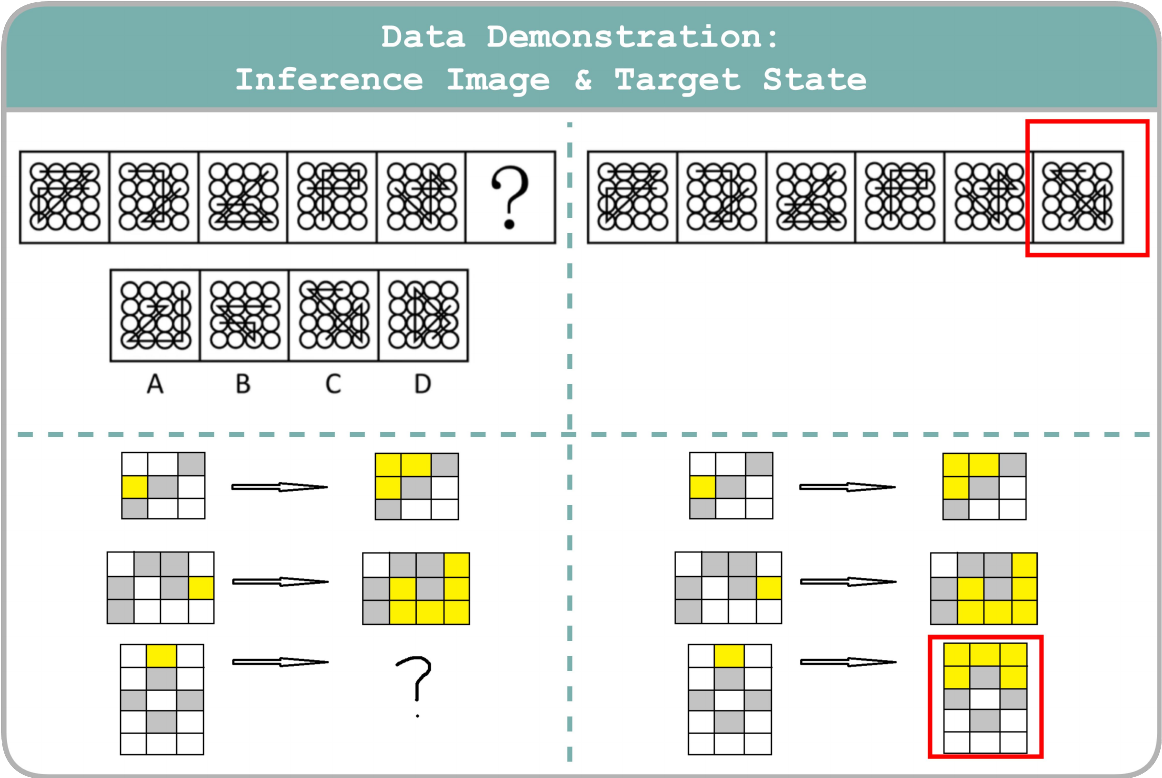}
  \vspace{-2.2em}
\end{wrapfigure}
The core idea behind these metrics is to focus on task-relevant regions within the visual input, rather than evaluating the entire frame. For each sample, we manually annotate the target state with a bounding box that specifies the area of interest. The cropped regions from the model's output and ground truth are passed through DINO to extract high-dimensional semantic features. Cosine similarity between these features quantifies alignment, with task-specific thresholds determining correctness. This approach ensures robustness to low-level variations while capturing high-level semantic alignment. DINO-based metrics provide a flexible framework for assessing diverse visual reasoning tasks, combining localized evaluation with powerful feature extraction to bridge the gap between pixel-level comparisons and semantic understanding.

\vspace{-0.4em}
\paragraph{DINO-X-based Grounding Metrics.} For tasks requiring complex visual grounding or dynamic target detection, we propose DINO-X-based \citep{ren2024dino} metrics, leveraging DINO-X's powerful grounding capability. These metrics are particularly suited for scenarios where target areas cannot be predefined or require advanced recognition, \textit{e.g.}, \textit{free-space mathematical reasoning}, \textit{object counting}, \textit{graph traversal}, and \textit{odd-one-out detection} tasks.

% \begin{figure}[!h]
% \centering
% \vspace{-0.8em}
% % \hspace*{0.42cm}
% \includegraphics[width=0.54\linewidth]{figures/app_c_1_8.pdf}
% \vspace{-2.2em}
% \end{figure}

The core idea is to dynamically ground task-relevant objects or regions based on high-level semantic prompts. For instance, in graph traversal tasks, we evaluate the number and types of nodes by grounding their visual
\begin{wrapfigure}{r}{0.48\textwidth}
\vspace{-0.2em}
 \centering
 \includegraphics[width=\linewidth]{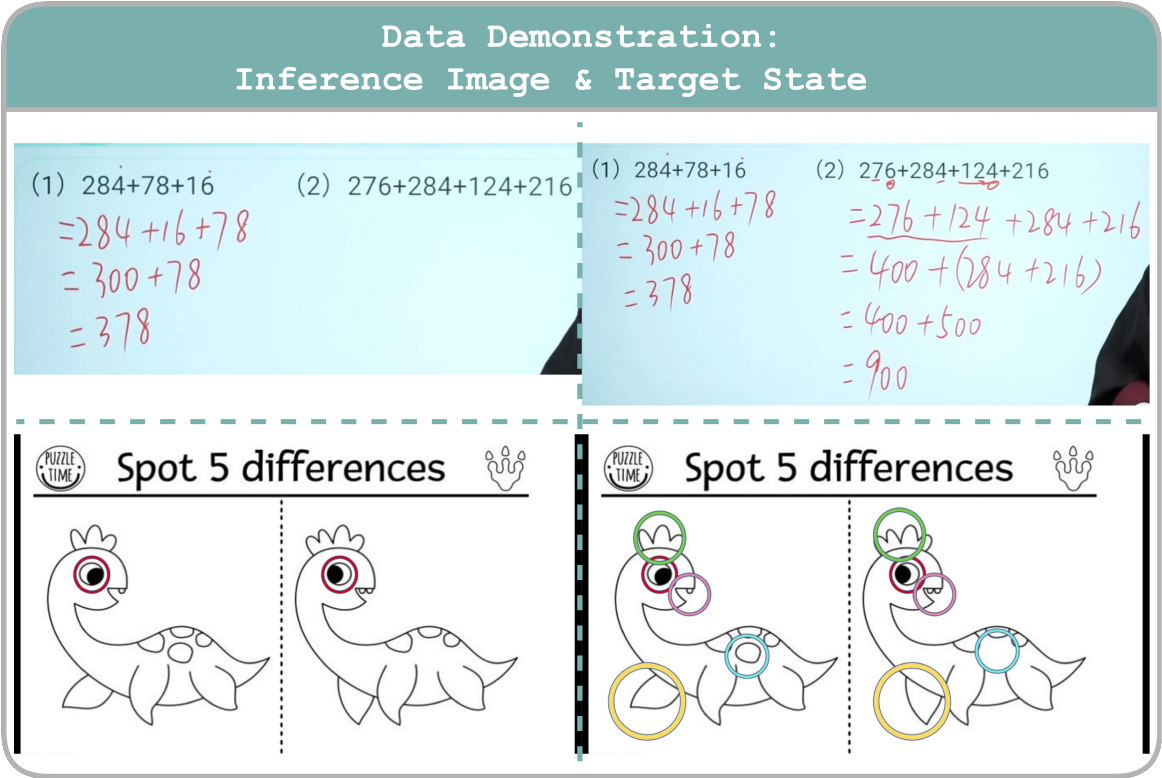}
  \vspace{-3.2em}
\end{wrapfigure}
attributes; and for odd-one-out tasks, we assess the positional and semantic differences of grounded objects (\textit{e.g.}, "colored circles") between the generated and ground truth frames. DINO-X enables flexible and robust evaluation by dynamically adapting to task-specific prompts and extracting high-level semantic features. This approach ensures that tasks with diverse visual reasoning requirements are evaluated consistently and accurately, even under challenging conditions where predefined regions or static rules are insufficient.

\subsection{Process-and-Goal Consistency}

\paragraph{DINO-X-based Tracking Metrics.} While final-state validation is sufficient for some tasks, many require evaluating the entire process to ensure both the correctness of the goal and the validity of the intermediate steps. To address this, we propose DINO-X-based tracking metrics that leverage video tracking and trajectory analysis to assess process-and-goal consistency. These metrics are particularly suitable for tasks such as maze solving, where the solution must avoid invalid actions (\textit{e.g.}, crossing walls or boundaries), and sequential elimination tasks, where objects must disappear in a specific order.

% \begin{figure}[!h]
% \centering
% \vspace{-0.8em}
% % \hspace*{0.42cm}
% \includegraphics[width=0.54\linewidth]{figures/app_c_2_1.pdf}
% \vspace{-2.2em}
% \end{figure}

\begin{wrapfigure}{r}{0.54\textwidth}
\vspace{-1.2em}
 \centering
 \includegraphics[width=\linewidth]{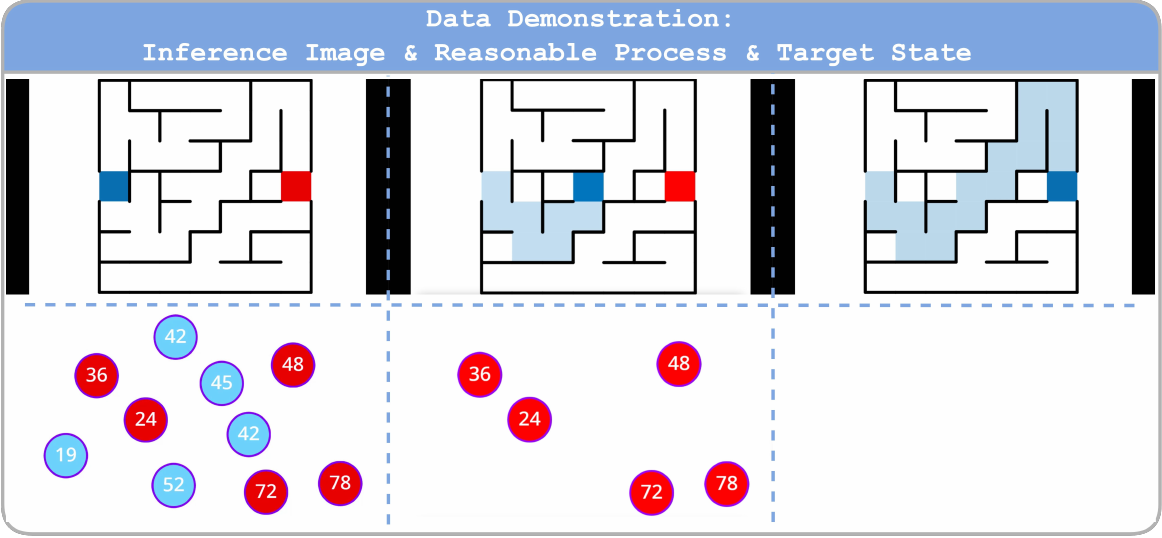}
  \vspace{-2.2em}
\end{wrapfigure}
The core methodology involves using DINO-X's visual grounding capabilities to track task-relevant objects or regions across frames. For example, in trajectory-based tasks, we extract object trajectories by uniformly sampling frames and grounding specific prompts (\textit{e.g.}, "blue block") to detect and record object positions over time. Trajectories are then compared against ground truth, ensuring alignment in both spatial and temporal dimensions; and for sequential tasks, we analyze the presence and disappearance of objects (\textit{e.g.}, "blue ball", "red ball") across sampled frames. The metric validates both the final state (\textit{e.g.}, all objects are eliminated) and the intermediate process (\textit{e.g.}, objects disappear in the correct sequence).

\vspace{-0.4em}
\paragraph{Gemini-based QA Metrics.} For tasks requiring extensive factual reasoning, such as action planning or tool use, traditional metrics based on visual grounding or trajectory analysis may fall short in capturing the nuanced logical dependencies and causal relationships inherent to these tasks. To address this limitation, we introduce VLM-based QA Metrics \citep{zhang2024mathverse, lu2024mathvista, zhang2024lmms}, that leverages the reasoning capabilities of Gemini-2.5-Pro \citep{deepmind_gemini_pro_2024} to assess task performance through question answering. 

% \begin{figure}[!h]
% \centering
% \vspace{-0.8em}
% % \hspace*{0.42cm}
% \includegraphics[width=1\linewidth]{figures/app_c_2_2.pdf}
% \vspace{-2.2em}
% \end{figure}
\begin{wrapfigure}{r}{0.54\textwidth}
\vspace{-1.2em}
 \centering
 \includegraphics[width=\linewidth]{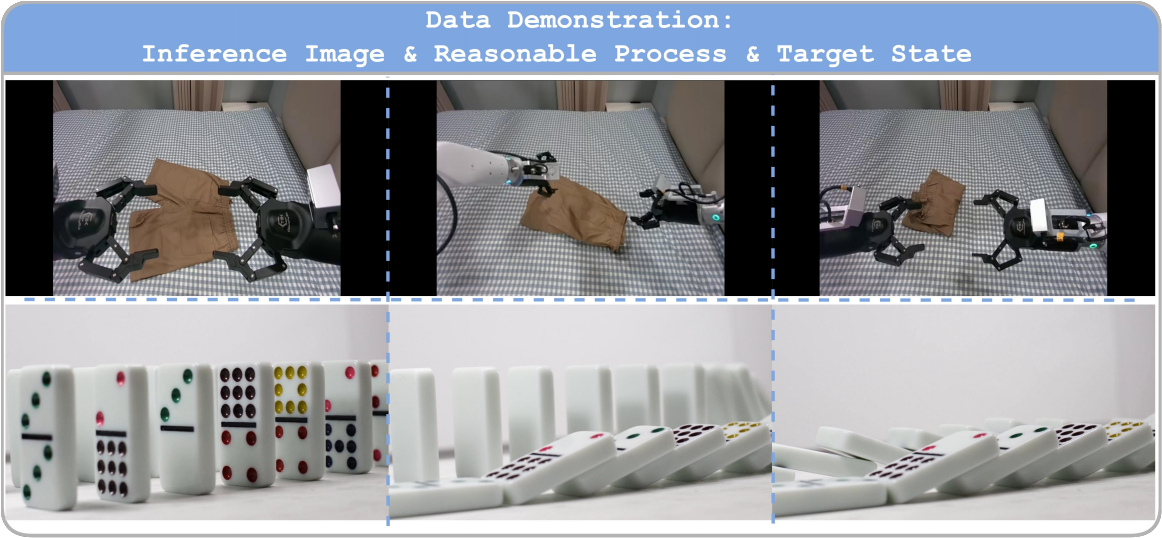}
  \vspace{-2.2em}
\end{wrapfigure}
Specifically, for each sample in this category, we design two or three binary questions tailored to the task's core requirements (\textit{e.g.}, \textit{"Is the wrench picked up in the video?"}). These questions are constructed to capture key aspects of the task's correctness, including intermediate actions, causal dependencies, and goal achievement. The generated video is then provided to Gemini-2.5-Pro along with the questions, and its responses are compared against the ground truth. A sample is deemed correct only if all three answers align with the ground truth, ensuring a high standard of evaluation fidelity.

\section{More Details of \ourmethod}
\label{app:tpo}

\subsection{Prompt Design}

Following TextGrad {\small\textcolor{gray}{[Nature'25]}}~\citep{yuksekgonul2025optimizing}, we adopt GPT-4o \citep{achiam2023gpt} as the optimizer and adopt the vanilla prompts for textual gradient calculation and prompt update from its implementation. To meet the requirements of video generation optimization, we further designed the textual loss calculation prompt:

\begin{tcolorbox}[notitle, sharp corners, breakable, colframe=BlueGreen!60, colback=gray!10, 
       boxrule=3pt, boxsep=0.5pt, enhanced, 
       shadow={3pt}{-3pt}{0pt}{opacity=1,newgray},
       title={Textual Loss Calculation}]\label{box:textual_loss}
       \footnotesize
       \setstretch{1}
       {\fontfamily{pcr}\selectfont
\begin{lstlisting}
"""You are a video generation system optimization expert tasked with evaluating a target text prompt and the generated video. Analyze the strengths and weaknesses of each generated video step by step, and explain why the video is not good or why it is good.

**Current Prompt**:
{current_prompt}

**Reasoning Task**:
{task_definition}

**Note**:
- The videos were stitched together vertically to form a single video for comparison purposes.
- Your output should only include the analysis.
- There may be instances where both videos are subpar, necessitating strict adherence to the task definition.

**Input Videos**:
{input_videos}
"""

\end{lstlisting}
}
\end{tcolorbox}

\begin{figure*}[!b]
\centering
% \vspace{-0.4em}
\includegraphics[width=\linewidth]{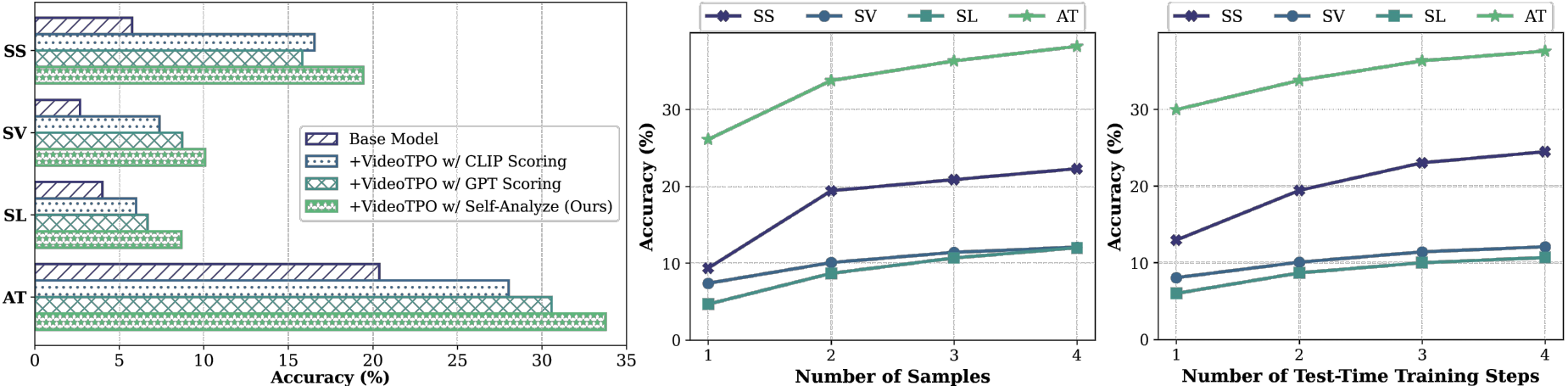}
\vspace{-1.9em}
 \caption{(\textbf{\textit{Left}}) Analysis of \ourmethod's rewarding strategies; (\textbf{\textit{Middle}}) Scaling width across sample numbers; and (\textbf{\textit{Right}}) Scaling depth across test-time training steps. }
\label{fig:app_d_2}
\vspace{-0.6em}
\end{figure*}

\subsection{More Analysis}

\paragraph{Analysis of Self-Analysis \textit{vs.} Reward Model.}

In TPO's \citep{li2025testtime} setting, a reward model is employed to select a preferred sample and a non-preferred sample from the generated candidates, which are then used to compute textual loss and gradients. However, \ourmethod eliminates the need for an additional reward model by leveraging task-specific VLMs (\textit{i.e.}, GPT-4o) to conduct self-analysis among candidate samples. The self-analysis process identifies strengths and weaknesses of each sample, directly informing optimization without relying on external scoring models.

To validate the effectiveness of our strategy, we compare \ourmethod with two widely-used reward strategies: CLIP scoring and GPT scoring, as shown in Figure~\ref{fig:app_d_2} (\textit{Left}). Results show that \ourmethod achieves significantly better accuracy, outperforming these reward-based methods across all reasoning dimensions. This advantage is likely due to the complexity of reasoning tasks, where candidate samples often exhibit subtle differences. In such cases, relying on a reward model to identify the "best" and "worst" samples provides limited utility, while self-analysis enables a more nuanced understanding of sample quality.

\paragraph{Analysis of Scaling in Width and Depth.}

To further evaluate the scalability of \ourmethod, we explore its performance under varying candidate sample numbers (\textit{width}) and scaling steps (\textit{depth}), with our default settings of $2$ samples and $2$ steps, respectively. Figure~\ref{fig:app_d_2} (\textit{Middle} and \textit{Right}) illustrates the impact of scaling in both dimensions.

In terms of \textit{width}, increasing the number of candidate samples consistently improves accuracy, as the self-analysis process benefits from a broader pool of options to identify optimal reasoning pathways. Similarly, scaling in \textit{depth}—by increasing the number of test-time training steps—also yields substantial performance gains, demonstrating the robustness of \ourmethod under extended optimization. These results highlight the flexibility and effectiveness of \ourmethod, making it a scalable solution for reasoning-intensive video generation tasks.

% \section{More Experimental Settings and Analysis}
% \label{app:analysis}

\section{Exhibition Board}
\label{app:exhibition}

\begin{figure*}[!t]
\centering
\vspace{-0.4em}
\includegraphics[width=0.98\linewidth]{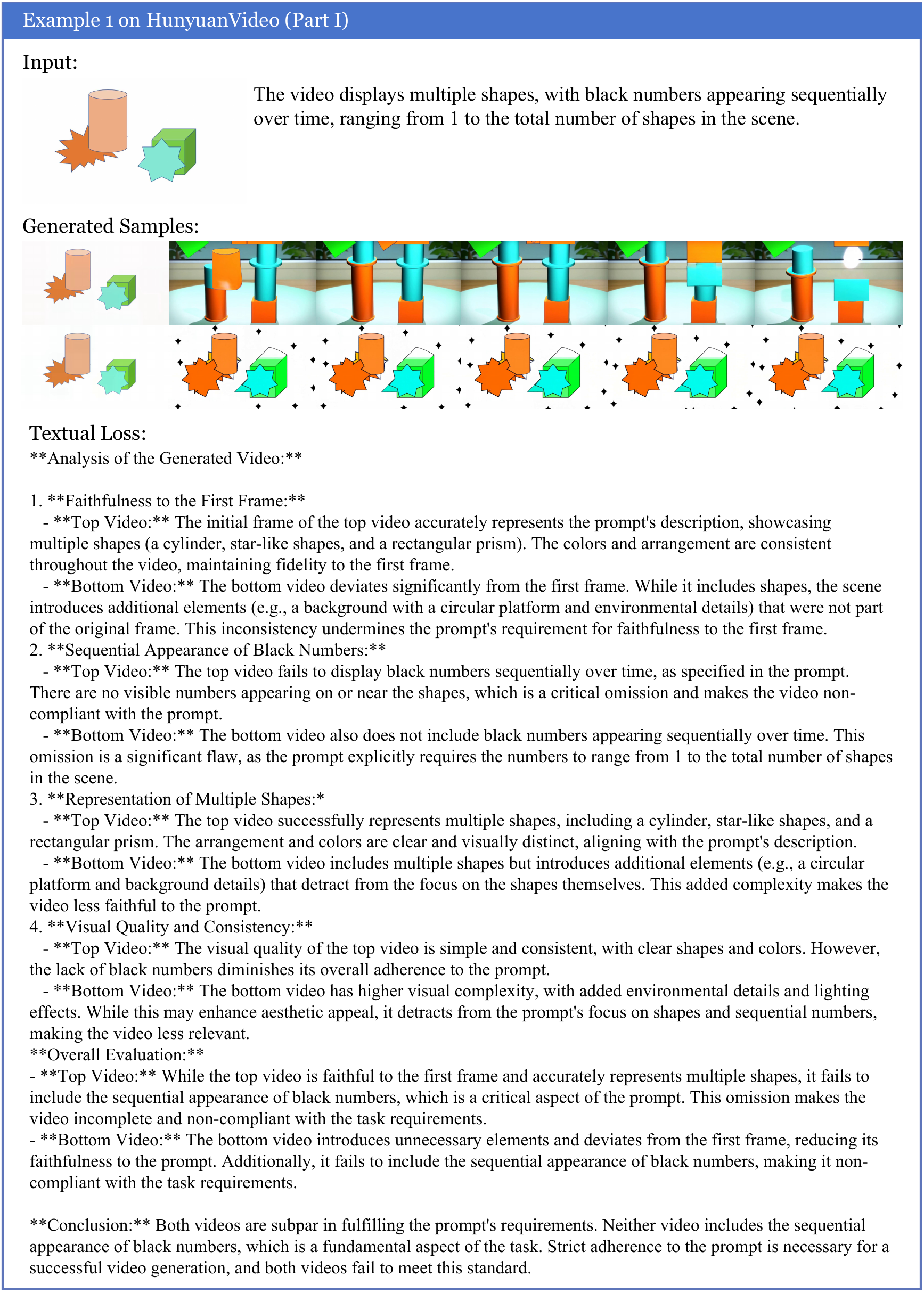}
\vspace{-0.6em}
 \caption{Case demonstration of \ourmethod. }
\label{fig:exhibition2-1}
% \vspace{-0.5em}
\end{figure*}

\begin{figure*}[!t]
\centering
\vspace{-0.4em}
\includegraphics[width=0.98\linewidth]{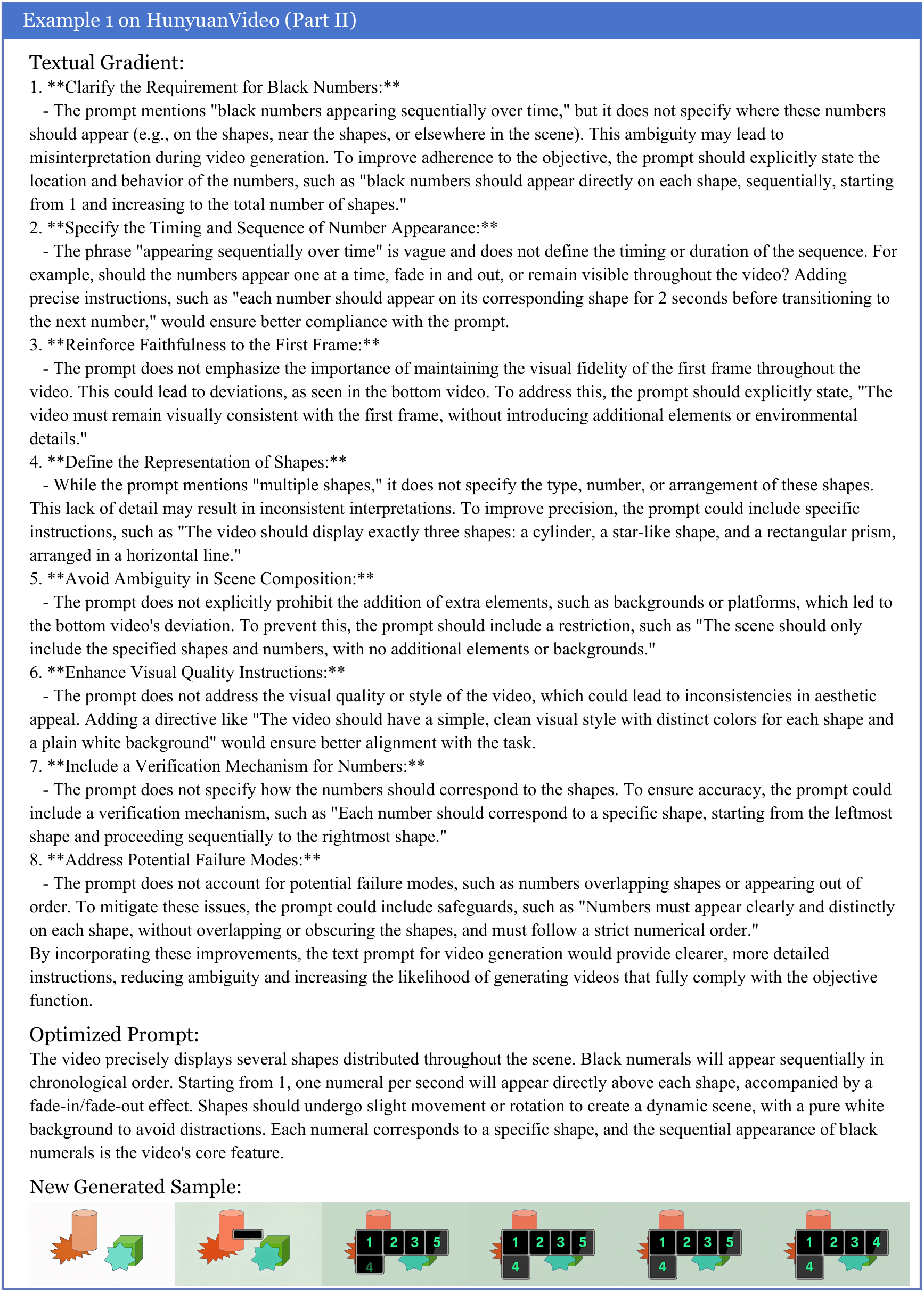}
\vspace{-0.6em}
 \caption{Case demonstration of \ourmethod. }
\label{fig:exhibition2-2}
% \vspace{-0.5em}
\end{figure*}

\begin{figure*}[!t]
\centering
\vspace{-0.4em}
\includegraphics[width=0.98\linewidth]{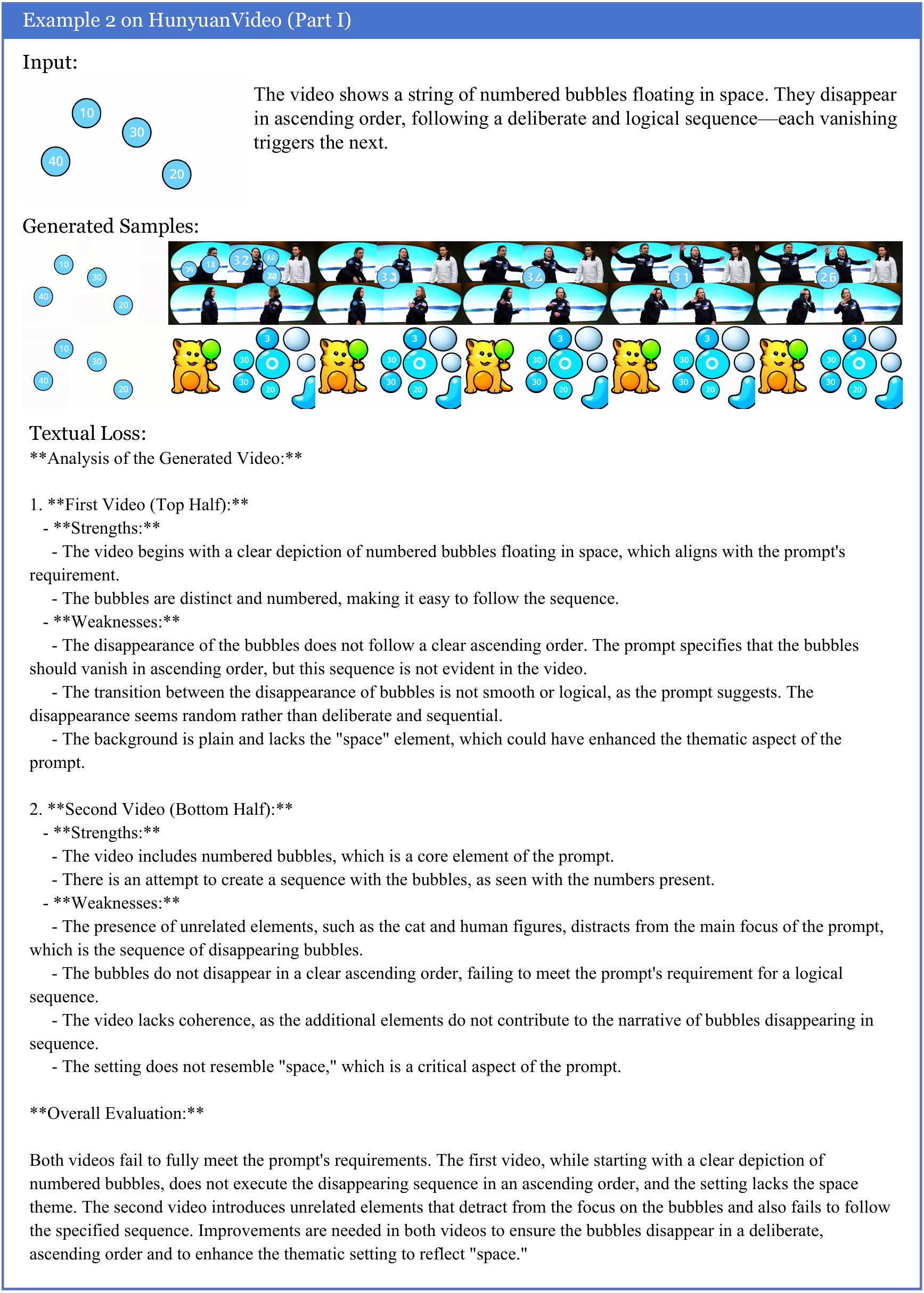}
\vspace{-0.6em}
 \caption{Case demonstration of \ourmethod. }
\label{fig:exhibition2-3}
% \vspace{-0.5em}
\end{figure*}

\begin{figure*}[!t]
\centering
\vspace{-0.4em}
\includegraphics[width=0.98\linewidth]{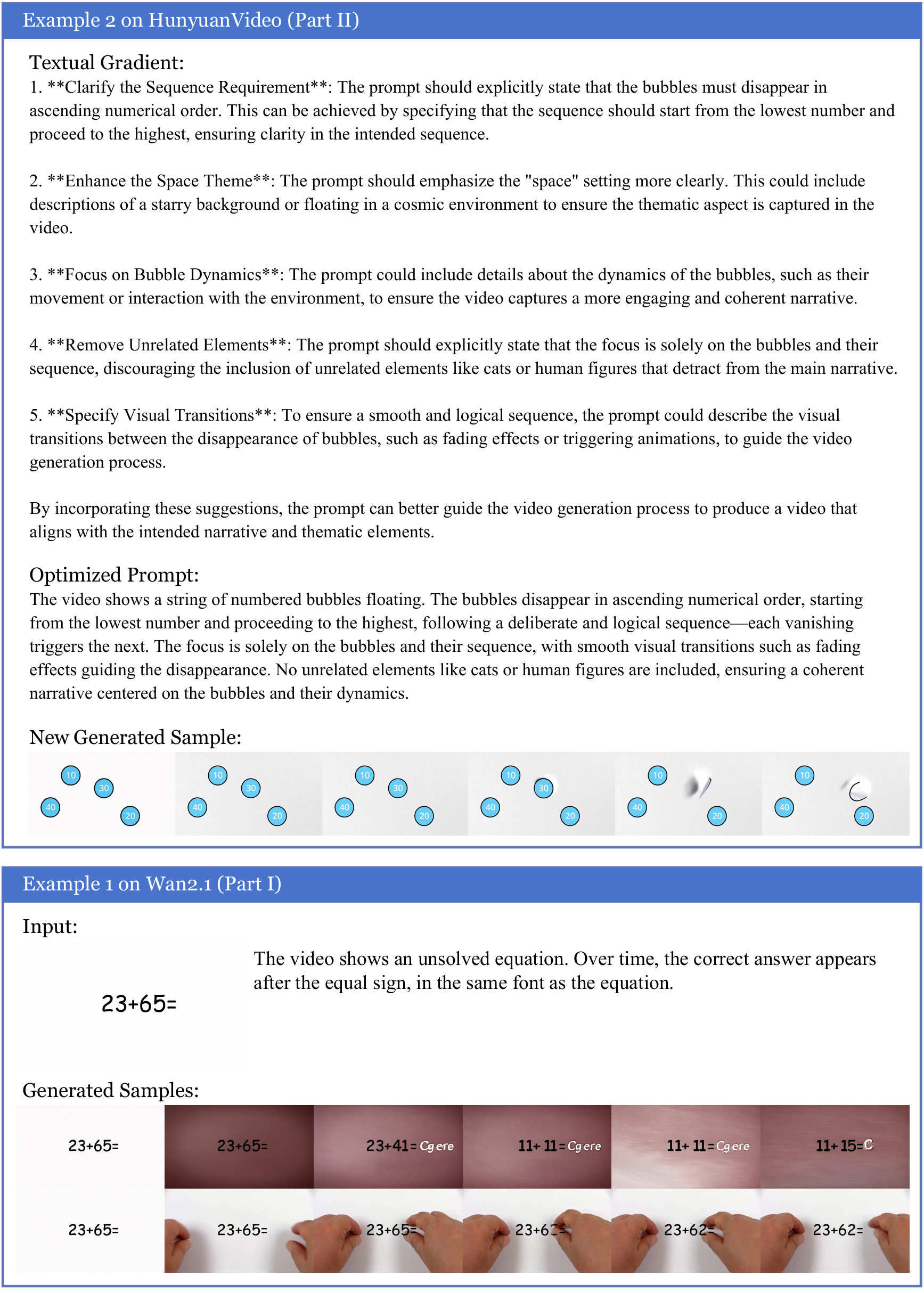}
\vspace{-0.6em}
 \caption{Case demonstration of \ourmethod. }
\label{fig:exhibition2-4}
% \vspace{-0.5em}
\end{figure*}

\begin{figure*}[!t]
\centering
\vspace{-0.4em}
\includegraphics[width=0.98\linewidth]{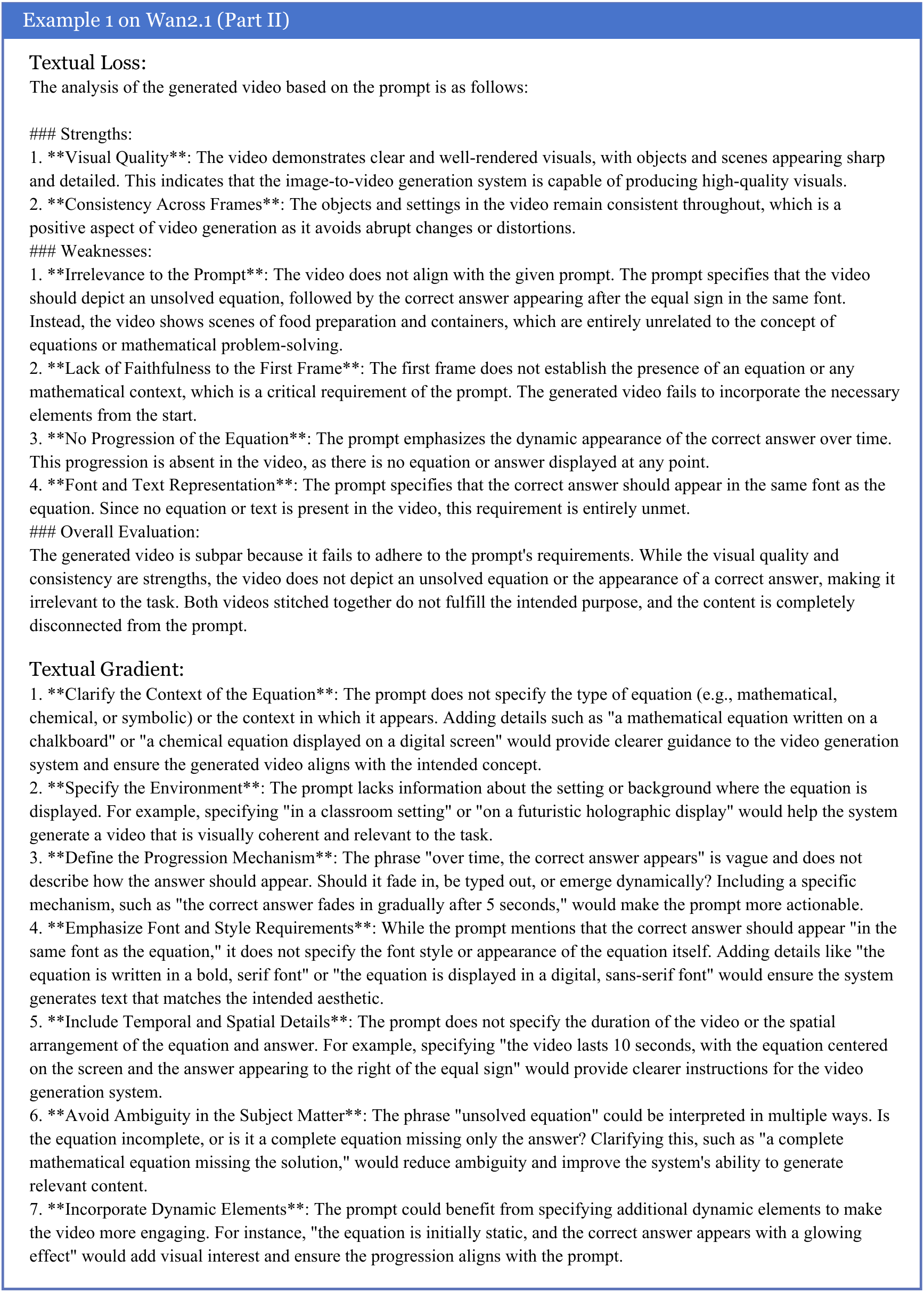}
\vspace{-0.6em}
 \caption{Case demonstration of \ourmethod. }
\label{fig:exhibition2-5}
% \vspace{-0.5em}
\end{figure*}

\begin{figure*}[!t]
\centering
\vspace{-0.4em}
\includegraphics[width=0.98\linewidth]{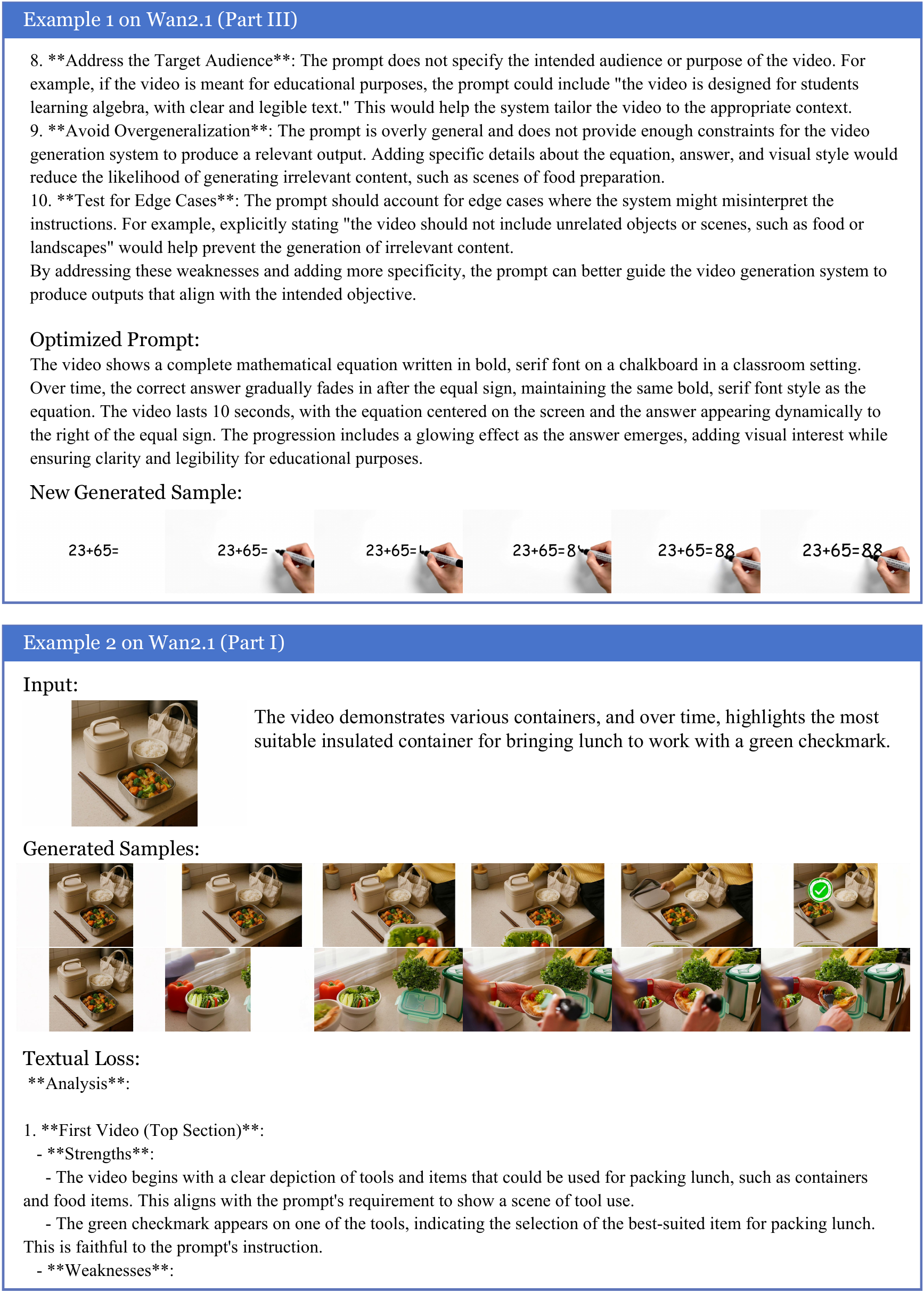}
\vspace{-0.6em}
 \caption{Case demonstration of \ourmethod. }
\label{fig:exhibition2-6}
% \vspace{-0.5em}
\end{figure*}

\begin{figure*}[!t]
\centering
\vspace{-0.4em}
\includegraphics[width=0.98\linewidth]{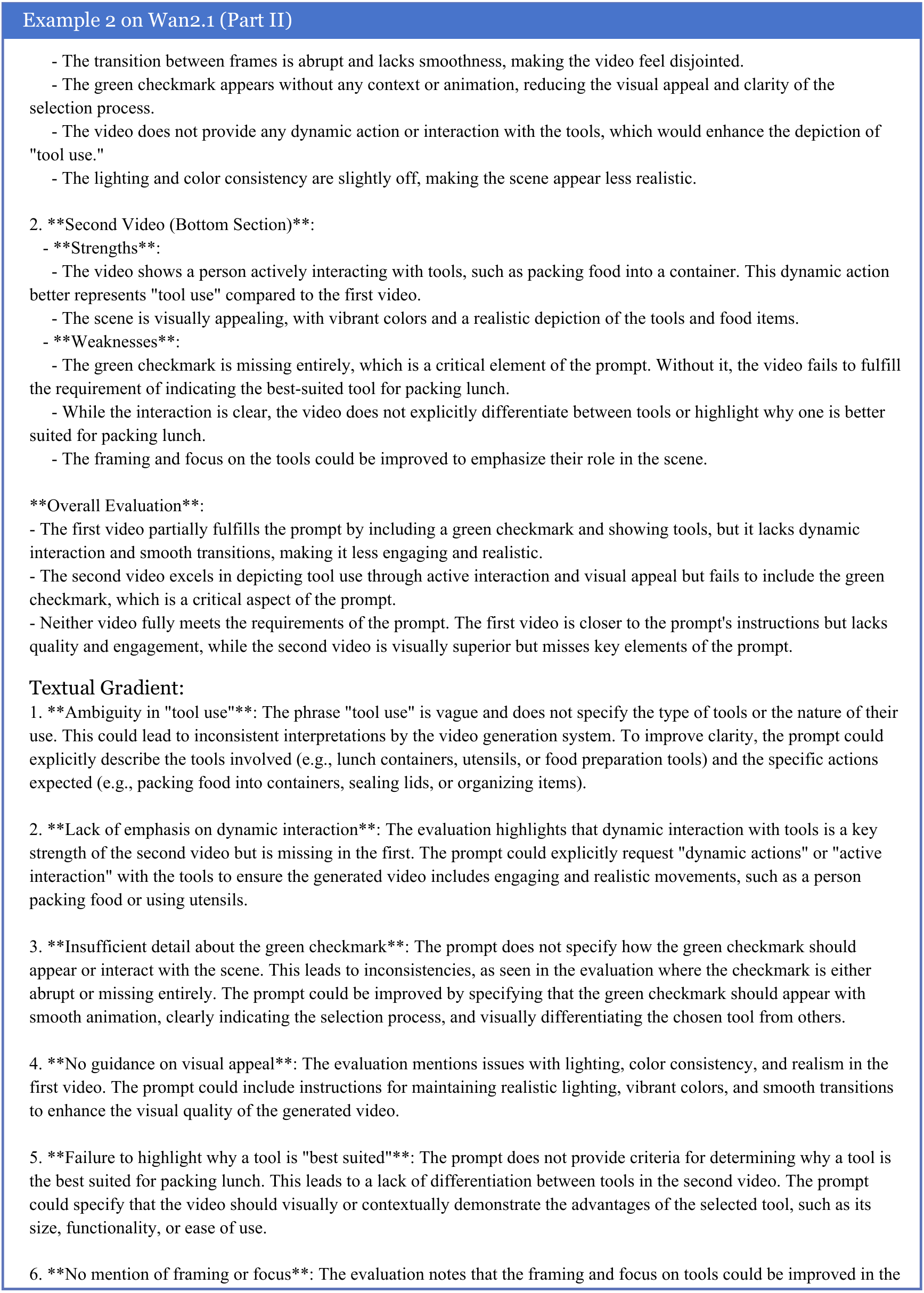}
\vspace{-0.6em}
 \caption{Case demonstration of \ourmethod. }
\label{fig:exhibition2-7}
% \vspace{-0.5em}
\end{figure*}

\begin{figure*}[!t]
\centering
\vspace{-0.4em}
\includegraphics[width=0.98\linewidth]{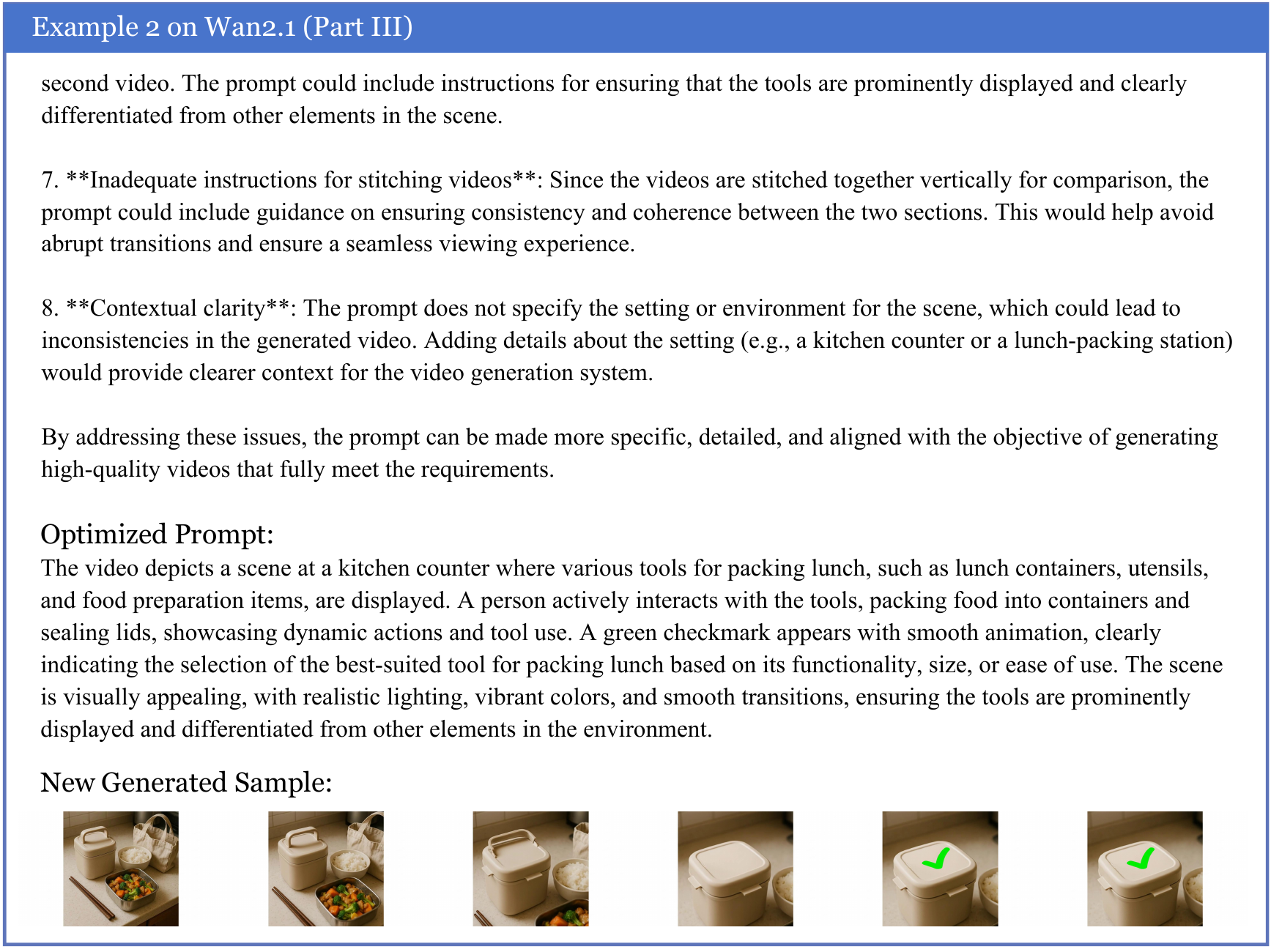}
\vspace{-0.6em}
 \caption{Case demonstration of \ourmethod. }
\label{fig:exhibition2-8}
% \vspace{-0.5em}
\end{figure*}

\paragraph{Demonstration of Results with \ourmethod.} Here we provide qualitative results of \ourmethod from Figure~\ref{fig:exhibition2-1} to Figure~\ref{fig:exhibition2-8}.

\vspace{-1em}
\paragraph{Demonstration of Results on \ourbench.}
% We further demonstrate more comparison results of evaluation on \ourbench from Figure~\ref{fig:exhibition1-1} to Figure~\ref{fig:exhibition1-12}.
For more comparison results of evaluation on \ourbench, please refer to our project page: \url{https://haroldchen19.github.io/TiViBench-Page/}.

\end{document}